\documentclass[10pt,twocolumn,letterpaper]{article}

\usepackage[pagenumbers]{cvpr}

\usepackage[dvipsnames]{xcolor}

\newcommand{\ar}[1]{\textsc{#1}}

\definecolor{cvprblue}{rgb}{0.21,0.49,0.74}
\usepackage[pagebackref,breaklinks,colorlinks,citecolor=cvprblue]{hyperref}

\usepackage{amsmath}
\usepackage{amssymb}
\usepackage{booktabs}
\usepackage{bm}

\usepackage[whole]{bxcjkjatype}
\usepackage{adjustbox}
\usepackage{multirow}
\usepackage{mathtools}
\usepackage{capt-of}
\usepackage{cuted}
\usepackage{xfrac}
\usepackage{microtype}
\usepackage{colortbl}
\usepackage{xcolor}
\usepackage{makecell}
\usepackage{pifont}
\usepackage{transparent}
\usepackage{float}
\usepackage{caption}
\usepackage{subcaption}
\usepackage{graphicx}

\makeatletter
\DeclareRobustCommand\onedot{\futurelet\@let@token\@onedot}
\def\@onedot{\ifx\@let@token.\else.\null\fi\xspace}
\def\ie{\emph{i.e}\onedot}

\makeatother

\renewcommand{\paragraph}{\vspace{1mm}\noindent\textbf}

\title{
\vspace{-7.5mm}Retrieval-Augmented Layout Transformer \\for Content-Aware Layout Generation}

\author{
    Daichi Horita\textsuperscript{1} \quad Naoto Inoue\textsuperscript{2} \quad Kotaro Kikuchi\textsuperscript{2} \quad Kota Yamaguchi\textsuperscript{2} \quad Kiyoharu Aizawa\textsuperscript{1} \\
$^{1}$The University of Tokyo \quad ${^2}$CyberAgent \\
    {\tt\small \{horita,aizawa\}@hal.t.u-tokyo.ac.jp \quad}
    \\ 
    {\tt\small 
    \{inoue\_naoto,kikuchi\_kotaro\_xa,yamaguchi\_kota\}@cyberagent.co.jp}
}

\definecolor{g}{gray}{0.82}
\definecolor{gg}{gray}{0.82}
\definecolor{redd}{rgb}{1.0 0.29 0}
\definecolor{blue}{rgb}{0.0, 0.35, 1.0}
\definecolor{greenn}{rgb}{0.011, 0.686, 0.478}
\definecolor{cadmiumgreen}{rgb}{0.0, 0.42, 0.24}
\definecolor{ceruleanblue}{rgb}{0.16, 0.32, 0.75}

\newcommand{\gb}[1]{\multirow{-1}{*}{\cellcolor{g}\makecell[c]{#1}}}
\newcommand{\tp}[1]{\transparent{0.5}{#1}}

\newcommand{\D}[1]{\textbf{#1}}
\newcommand{\E}[1]{#1}

\newcommand{\F}[1]{\multirow{-1}{*}{\cellcolor{g}\makecell[c]{#1}}}

\newcommand{\tabmain}{
  \begin{table*}[t]
    \centering
\resizebox{0.9 \linewidth}{!}{

\begin{tabular}{@{}l c c c c c c c c c c c@{}}
\toprule
\multirow{3}{*}{Method}& \multirow{3}{*}{\#Params}  & \multicolumn{5}{c}{PKU} & \multicolumn{5}{c}{CGL} \\
\cmidrule(lr){3-7} \cmidrule(lr){8-12}
~ & ~ &
\multicolumn{2}{c}{Content} &\multicolumn{3}{c}{Graphic} & 
\multicolumn{2}{c}{Content} & \multicolumn{3}{c}{Graphic} \\
\cmidrule(lr){3-4} \cmidrule(lr){5-7}
\cmidrule(lr){8-9} \cmidrule(lr){10-12}
~ & ~ 
& Occ $\downarrow$ & Rea $\downarrow$ & Und $\uparrow$ & Ove $\downarrow$ & FID$\downarrow$
& Occ $\downarrow$ & Rea $\downarrow$ & Und $\uparrow$ & Ove $\downarrow$ & FID$\downarrow$ \\
\midrule

\tp{Real Data} & -
& \tp{0.112} & \tp{0.0102} & \tp{0.99} & \tp{0.0009} & \tp{1.58} 
& \tp{0.125} & \tp{0.0170} & \tp{0.98} & \tp{0.0002} & \tp{0.79}
\\

\tp{Top-1 Retrieval} & -
& \tp{0.212} & \tp{0.0218} & \tp{0.99} & \tp{0.002} & \tp{1.43} 
& \tp{0.214} & \tp{0.0266} & \tp{0.99} & \tp{0.0005} & \tp{0.93}
\\

CGL-GAN~\cite{cglgan} & 41M
 & 0.138 & 0.0164 & 0.41 & 0.074& 34.51
& 0.157 & 0.0237 & 0.29 & 0.161 & 66.75
\\

DS-GAN~\cite{pkuposter} & 30M
& 0.142 & 0.0169 & \E{0.63} & 0.027 & \E{11.80}
& 0.141 & 0.0229 & 0.45 & 0.057 & 41.57
\\

ICVT~\cite{cao2022geometry} & 50M & 0.146 & 0.0185 & 0.49 & 0.318 & 39.13 & \D{0.124} & 0.0205 & 0.42 & 0.310 & 65.34
\\

LayoutDM$^{\dagger}$~\cite{inoue2023layout} & 43M
& 0.150 & 0.0192 & 0.41 & 0.190 & 27.09
& \E{0.127} & 0.0192 & 0.82 & 0.020 & \E{2.36}
\\
Autoreg Baseline & 41M
 & 0.134 & 0.0164 & 0.43 & 0.019 & 13.59
 & 0.125 & \E{0.0190} & \E{0.92} & \E{0.011} & 2.89
\\

\gb{RALF (Ours)} & \gb{43M}
 & \gb{\D{0.119}} & \gb{\D{0.0128}} & \gb{\D{0.92}} & \gb{\D{0.008}} & \gb{\D{3.45}} 
 & \gb{0.125} & \gb{\D{0.0180}} & \gb{\D{0.98}} & \gb{\D{0.004}} & \gb{\D{1.32}}
\\

\bottomrule
				\end{tabular}
    }
    \caption{
      Unconstrained generation results on the PKU and CGL test split.
      Our RALF outperforms the Autoreg Baseline and achieves the best score on almost all metrics. 
      For reference, we show the Real Data and the Top-1 Retrieval baselines, which do not have a generator.
    }
    \label{tab:uncondtable}
  \end{table*}
}

\newcommand{\tabUncondWithNoAnnoData}{
  \begin{table}[t]
    \centering
\resizebox{1.0\linewidth}{!}{

\begin{tabular}{@{}l c c c c c c c c c @{}}
\toprule
  \multirow{3}{*}{Method} & \multicolumn{4}{c}{PKU unannotated} & \multicolumn{4}{c}{CGL unannotated} \\
\cmidrule(lr){2-5} \cmidrule(lr){6-9}
~ &
\multicolumn{2}{c}{Content} & \multicolumn{2}{c}{Graphic} &
\multicolumn{2}{c}{Content} & \multicolumn{2}{c}{Graphic} \\
\cmidrule(lr){2-3} \cmidrule(lr){4-5}
\cmidrule(lr){6-7} \cmidrule(lr){8-9}
~
& Occ $\downarrow$ & Rea $\downarrow$ & Und $\uparrow$ & Ove $\downarrow$ 
& Occ $\downarrow$ & Rea $\downarrow$ & Und $\uparrow$ & Ove $\downarrow$
\\
\midrule

CGL-GAN
& 0.191 & 0.0312 & 0.32 & 0.069
& 0.481 & 0.0568 & 0.26 & 0.269
\\ 

DS-GAN
& 0.180 & 0.0301 & \E{0.52} & 0.026
& 0.435 & 0.0563 & 0.29 & 0.071
\\

ICVT
& 0.189 & 0.0317 & 0.48 & 0.292
& 0.446 & 0.0425 & 0.67 & 0.301 
\\

LayoutDM$^{\dagger}$
& 0.165 & 0.0285  & 0.38 & 0.201 
& 0.421 & 0.0506  & 0.49 & 0.069
\\

Autoreg Baseline
& 0.154 & 0.0274 & 0.35 & 0.022 
& 0.384 & 0.0427 & 0.76 & 0.058 
\\

\gb{RALF (Ours)}
& \gb{\D{0.133}} & \gb{\D{0.0231}} & \gb{\D{0.87}} & \gb{\D{0.018}}
& \gb{\D{0.336}} & \gb{\D{0.0397}} & \gb{\D{0.93}} & \gb{\D{0.027}} 
\\
\bottomrule
				\end{tabular}
    }
    \caption{
      Unconstrained generation results on the PKU and CGL unannotated test split, which is real data without inpainting artifacts.
    }
    \label{tab:uncond_with_no_anno_data}
  \end{table}
}

\newcommand{\figRetrievalComparisonPlot}{
\begin{figure}
    \centering
    \includegraphics[width=.62\linewidth]{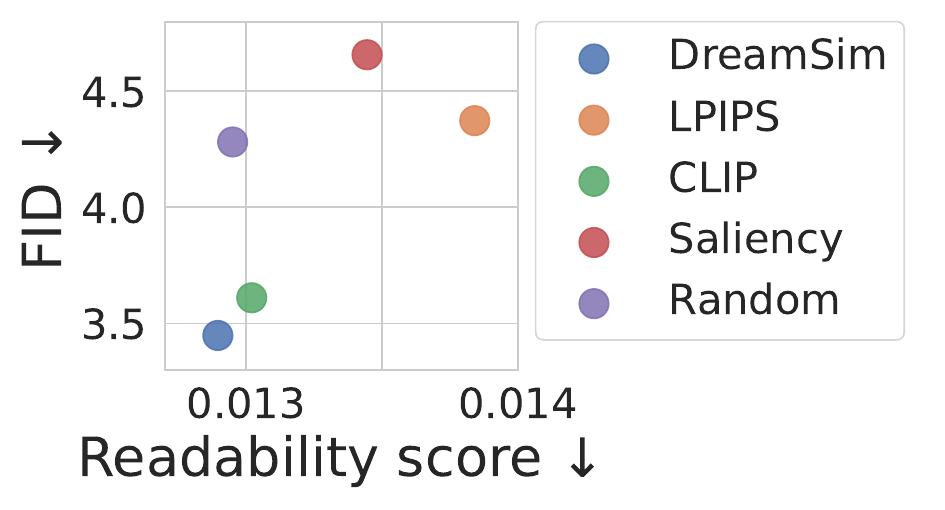}
    \vspace{-0.07cm}
    \caption{
    Comparison across different retrieval methods on the PKU test split.
    We report FID as the representative graphic metric and Readability score as the content metric.}
    \label{fig:retrievalComparisonPlot}
\end{figure}
}

\newcommand{\figRetrievalComparison}{
\setlength\tabcolsep{.5pt}
\begin{figure}
    \includegraphics[width=0.9\linewidth]{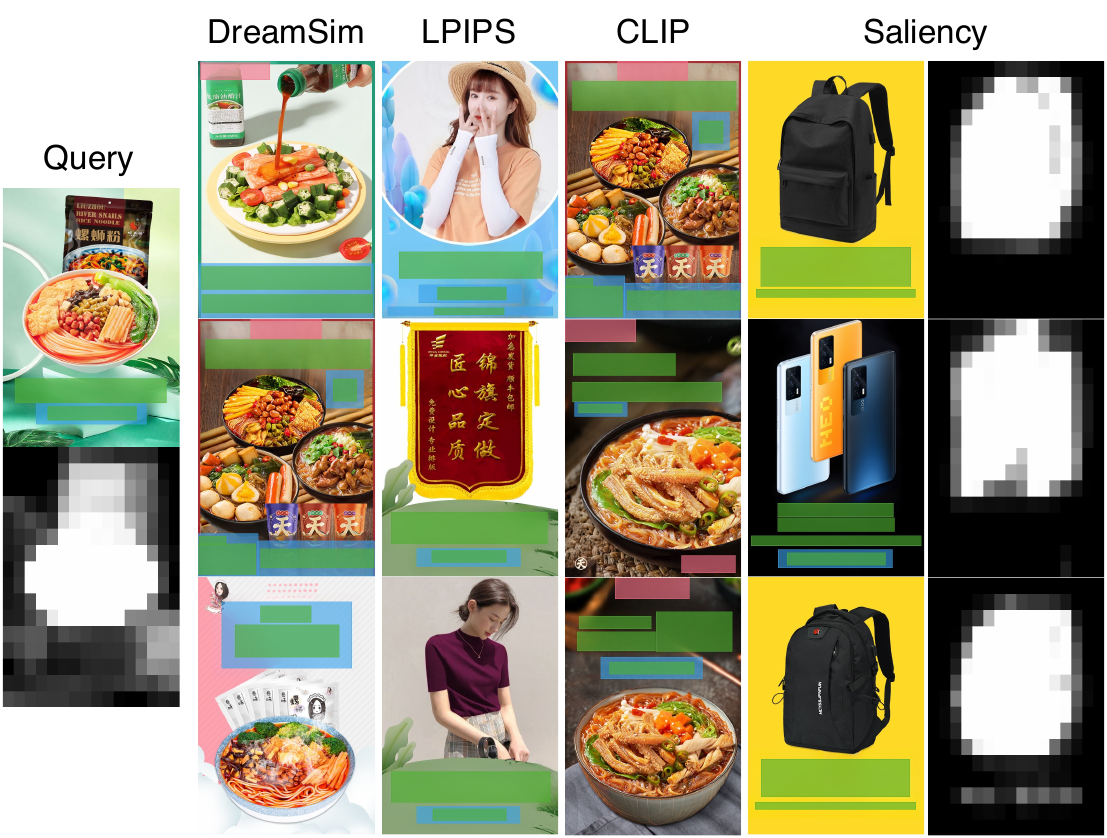}
    \centering
    \caption{
        Qualitative comparison of different retrieval methods.
        We show the query and the top-3 retrieved examples for each method.
    }
    \label{fig:retrievalComparison}
  \end{figure}
  \setlength\tabcolsep{6pt}
}

\newcommand{\tabNetworkParameter}{
  \begin{table}[t]
    \centering
    \begin{tabular}{@{}l c@{}}
\toprule
Module & \#Params \\
\midrule
    Image encoder (ResNet50)    & 25.02 M \\
    Image encoder (Trans Enc) & 4.74 M \\
    Constraint encoder          & 4.88 M   \\
    Retrieval augmentation           & 1.59 M   \\
    Layout decoder              & 6.59 M   \\
    \midrule
    Total                   & 42.82 M   \\

\bottomrule
    \end{tabular}
    \caption{
        The number of parameters of each module.
    }
    \label{tab:networkparameter}
  \end{table}
}

\newcommand{\tabAblationNetwork}{
  \begin{table}[t]
    \centering
    \resizebox{ \linewidth}{!}{
    \begin{tabular}{@{}l@{\hspace{2.0 mm}}l c c c c c@{}}
    \toprule
    ~ & Setting
    & Occ $\downarrow$ & Rea $\downarrow$ & Und $\uparrow$ & Ove $\downarrow$ & FID$\downarrow$ \\
    
\midrule
        \gb{\ar{a}} & \gb{Ours ($\mathrm{Concatenate}(f_\mathrm{I}, \tilde{f}_\mathrm{L}, f_\mathrm{C})$)}
         & \gb{\D{0.119}} & \gb{\D{0.0129}} & \gb{\underline{0.92}} & \gb{\underline{0.008}} & \gb{\underline{3.45}}
        \\
        \midrule
        & \textbf{What types of features to fuse?} \\
        \ar{b} & $\mathrm{Concatenate}(f_\mathrm{C}, )$
        & 0.134 & 0.0144 & \underline{0.92} & \underline{0.008} & 4.67
        \\
        \ar{c} & $\mathrm{Concatenate}(f_\mathrm{I}, \tilde{f}_\mathrm{L})$
        & 0.123 & \underline{0.0133} & 0.91 & \D{0.007} & 4.08
        \\
        \ar{d} &  $\mathrm{Concatenate}(f_\mathrm{I}, \tilde{f}_\mathrm{L}, f_\mathrm{C}, \tilde{f}_\mathrm{I})$
        & 0.141 & 0.0148 & \D{0.93} & 0.009  & 8.82
        \\
        \midrule
        & \textbf{Where to apply?} \\
        \ar{e} & Before Trans enc
         & \underline{0.120} & 0.0138 & 0.72 & 0.009 & \D{2.34}
         \\
\bottomrule
    \end{tabular}
    }
    \vspace{-0.2cm}
    \caption{
         Ablation study of RALF design on the PKU test split.
          The top two results are highlighted in \textbf{bold} and \underline{underline}, respectively.
          Features include the input canvas feature ($f_\mathrm{I}$), retrieved layouts feature ($\tilde{f}_\mathrm{L}$), cross-attended feature ($f_\mathrm{C}$), and retrieved images feature ($\tilde{f}_\mathrm{I}$).
          The full setting of our model (\ar{a}) is described in~\cref{eq:feature_fusion}.
    }
    \label{tab:ablation_network}
  \end{table}
}

\newcommand{\miniAblations}{

    \begin{figure}[t]
        \begin{minipage}[t]{0.475\linewidth}
            \centering
            \includegraphics[width=\linewidth]{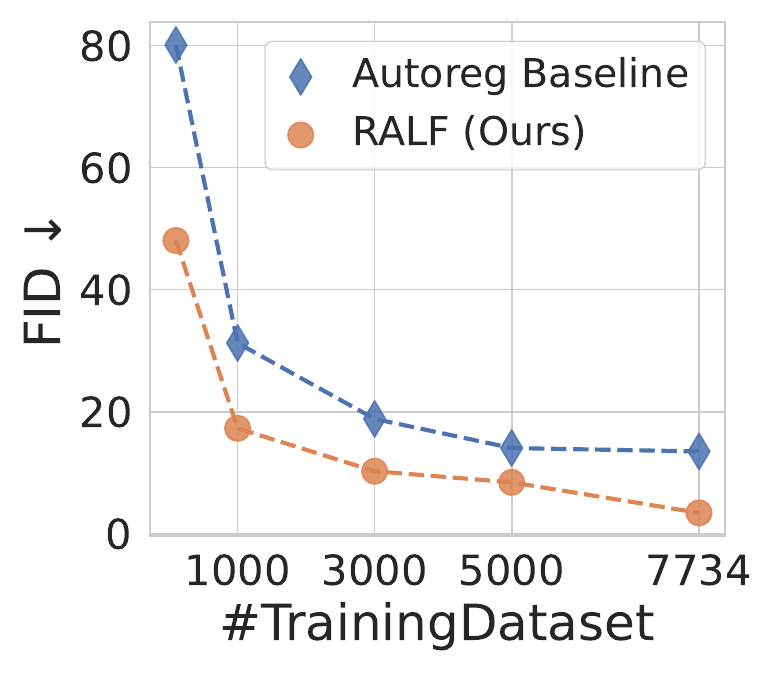}
            \captionof{figure}{
            FID over the training dataset size (\#TrainingDataset), which has up to 7,734 samples.
            }
            \label{minifig:ablation_training_sample}
        \end{minipage}%
        \hfill
        \begin{minipage}[t]{0.475\linewidth}
            \centering
            \includegraphics[width=\linewidth]{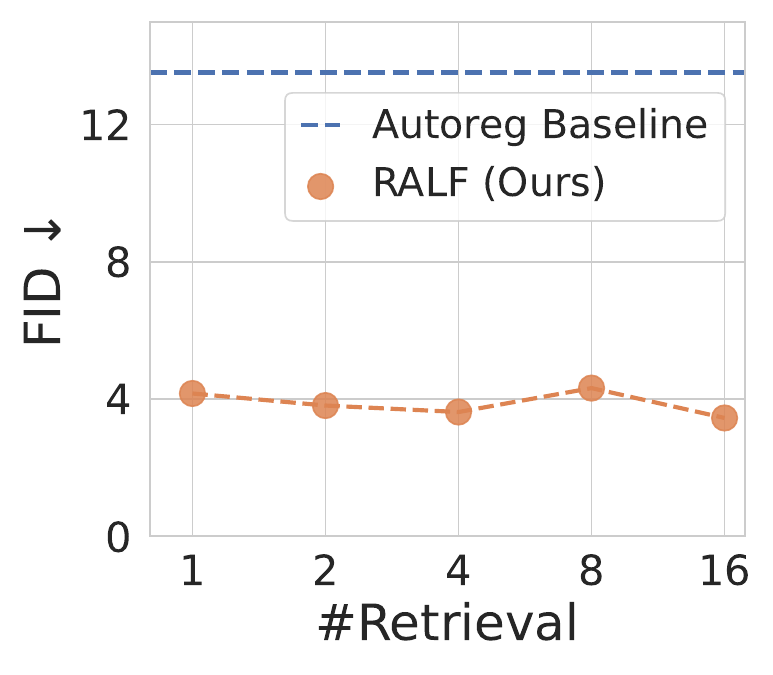}
            \captionof{figure}{
            FID over the retrieval size $K$ (\#Retrieval).
            }
            \label{minifig:ablation_k}
        \end{minipage}
    \end{figure}
}

\newcommand{\figFramework}{
\begin{figure*}
    \centering
    \includegraphics[width=0.95\linewidth]{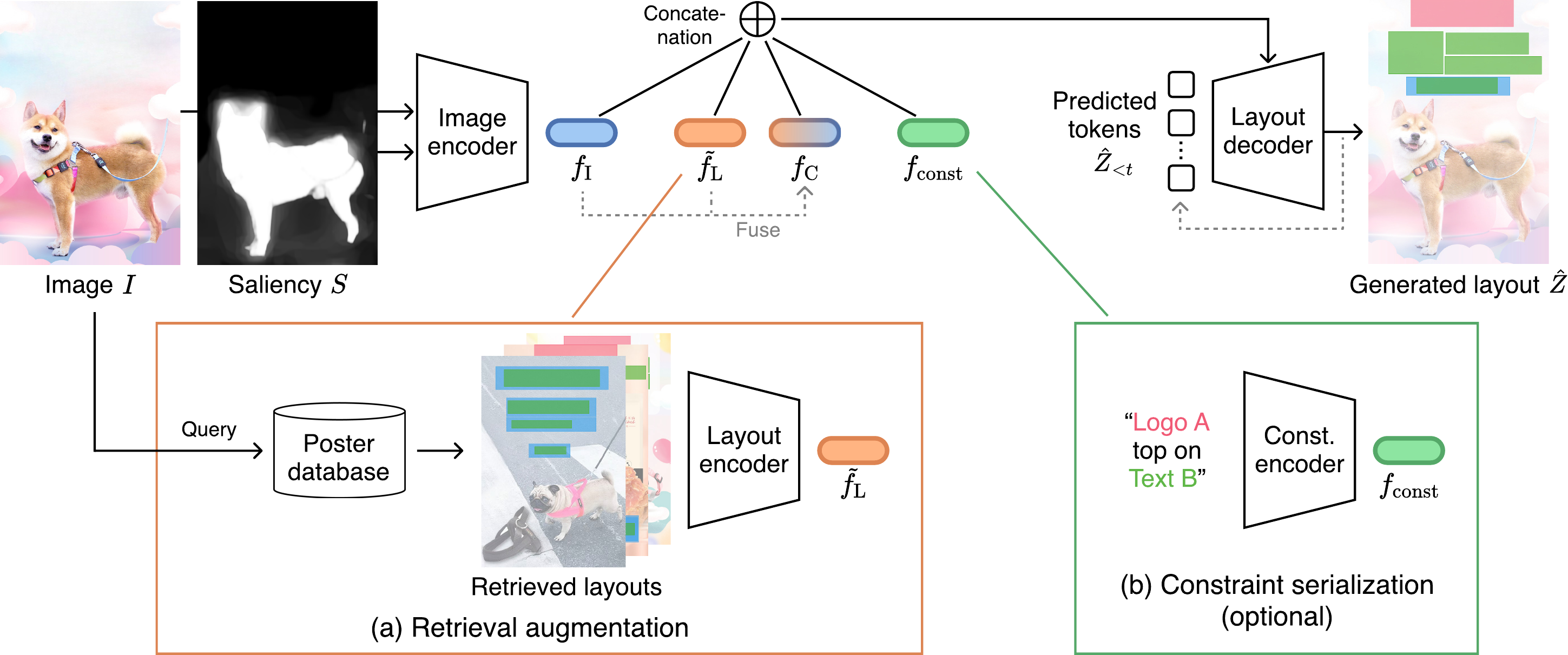}
    \caption{
        Overview of Retrieval-Augmented Layout Transformer (RALF).
        RALF takes a canvas image and a saliency map as input, and then autoregressively generates a layout along with the input image.
        Our model uses (a) retrieval augmentation that incorporates useful examples to better capture the relationship between the image and the layout, and (b) constraint serialization, an optional module that encodes user-specified requirements, enabling the generation of layouts that adhere to specific requirements for controllable generation.
    }
    \label{fig:framework}
  \end{figure*}
}

\newcommand{\tabAblationRA}{
  \begin{table}[t]
    \centering
    \resizebox{\linewidth}{!}{
    \begin{tabular}{@{}l c c c c c c@{}}
    \toprule
    Method & Retrieval
     & Occ $\downarrow$ & Rea $\downarrow$ &Und $\uparrow$ & Ove $\downarrow$ & FID$\downarrow$ \\
    \midrule
    CGL-GAN & ~
    & \D{0.138} & \D{0.0164} & 0.41 & 0.074 & 34.51 \\
    \gb{CGL-GAN} & \gb{\ding{51}}
    & \gb{\E{0.144}} & \gb{\D{0.0164}} & \gb{\D{0.63}} & \gb{\D{0.039}} & \gb{\D{13.28}}
    \\
    \midrule
    LayoutDM$^{\dagger}$ & ~
    & 0.150 & 0.0192 & 0.41 & 0.190 & 27.09 \\
    \gb{LayoutDM$^{\dagger}$} & \gb{\ding{51}}
    & \gb{\D{0.123}} & \gb{\D{0.0144}} & \gb{\D{0.51}} & \gb{\D{0.091}} & \gb{\D{10.03}} 
    \\
\bottomrule
    \end{tabular}
    }
    \caption{
         Retrieval augmentation for CGL-GAN and LayoutDM$^{\dagger}$ on the PKU test split.
    }
    \label{tab:ablation_ra}
  \end{table}
}

\newcommand{\tabCrossUncond}{
  \begin{table}[t]
    \centering
\resizebox{\linewidth}{!}{

\begin{tabular}{@{}c c c c c c c c c c c @{}}
\toprule
Train & Test & Method & Occ $\downarrow$ & Rea $\downarrow$ & Und $\uparrow$ & Ove $\downarrow$
\\
\midrule
\multirow{2}{*}{\makecell{CGL}} &
\multirow{2}{*}{\makecell{PKU}} &
Autoreg Baseline
& 0.176 & 0.0276 & 0.84 & 0.037
\\ 
~&~& \gb{RALF (Ours)}
& \gb{\D{0.144}} & \gb{\D{0.0249}} & \gb{\D{0.96}} & \gb{\D{0.023}}
\\
\midrule
\multirow{2}{*}{\makecell{PKU}} &
\multirow{2}{*}{\makecell{CGL}} &
Autoreg Baseline
& 0.341 & 0.0464 & 0.29 & 0.037 
\\ 
~&~& \gb{RALF (Ours)}
& \gb{\D{0.286}} & \gb{\D{0.0355}} & \gb{\D{0.79}} & \gb{\D{0.036}}
\\
\bottomrule
				\end{tabular}
    }
    \caption{
    Generation across the unannotated test splits.
    We train a model on PKU and then test it on CGL with the layout database of PKU, or vice versa.
    }
    \label{tab:cross_uncond}
  \end{table}
}

\newcommand{\teaser}{
\begin{figure}
    \centering
    \includegraphics[width=\linewidth]{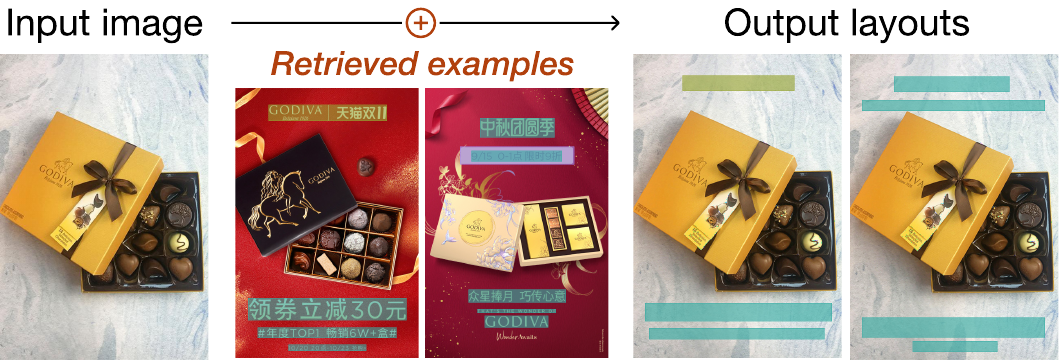}
    \caption{
        Retrieval-augmented content-aware layout generation.
        We retrieve nearest neighbor examples based on the input image and use them as a reference to augment the generation process.
    }
    \label{fig:teaser}
  \end{figure}
}

\newcommand{\figPKUTeiseiUncondSingleClumn}{
\begin{figure}
    \centering
    \includegraphics[width=\linewidth]{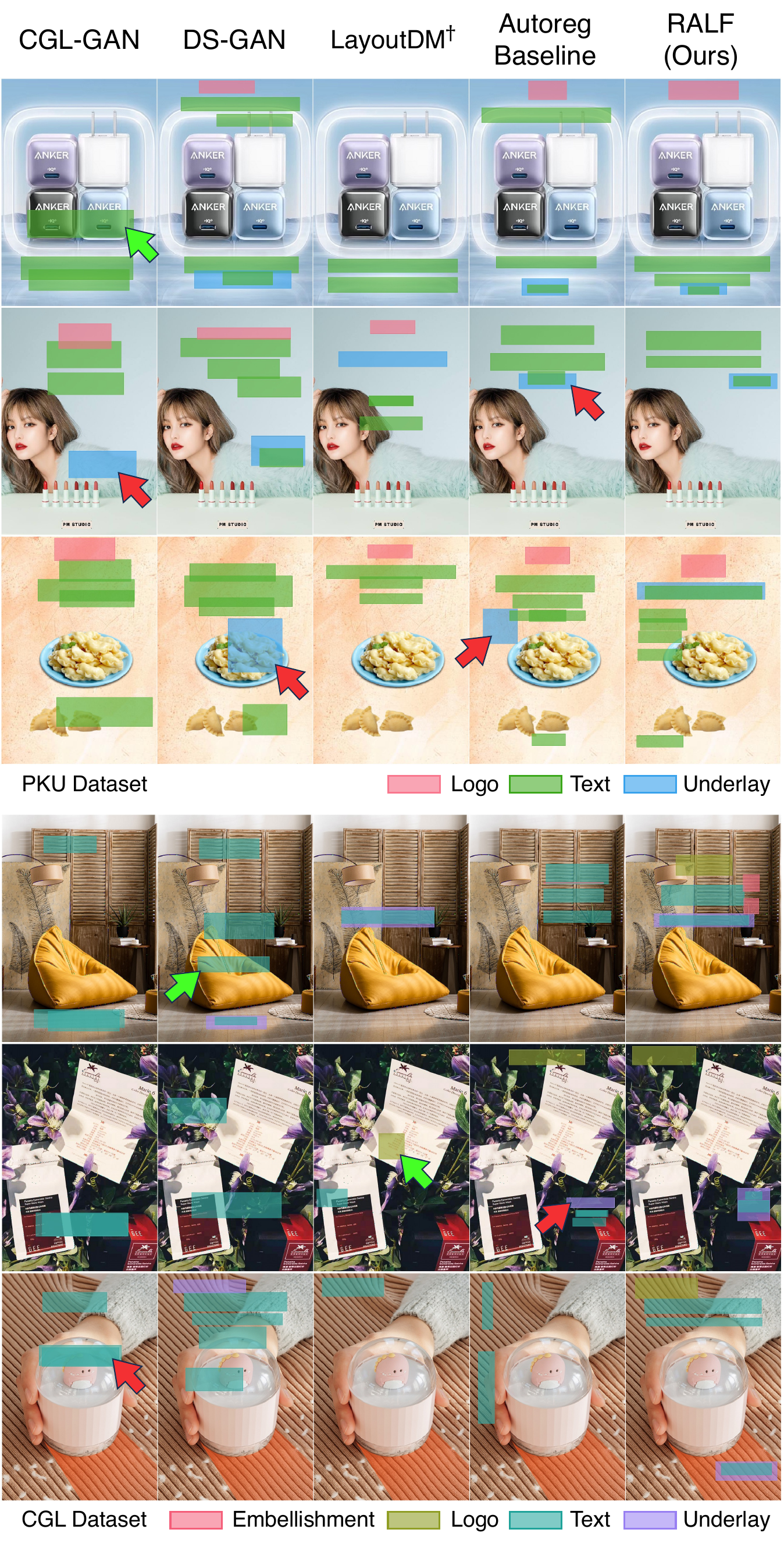}
    \caption{
        Visual comparison of unconstrained generation with baselines.
        Input canvases are selected from the unannotated split.
    }
    \label{fig:main_comparison}
  \end{figure}
}

\newcommand{\figRetrievalSizeQualitative}{
\begin{figure}
    \setlength\tabcolsep{.5pt}
    \centering
    \includegraphics[width=\linewidth]{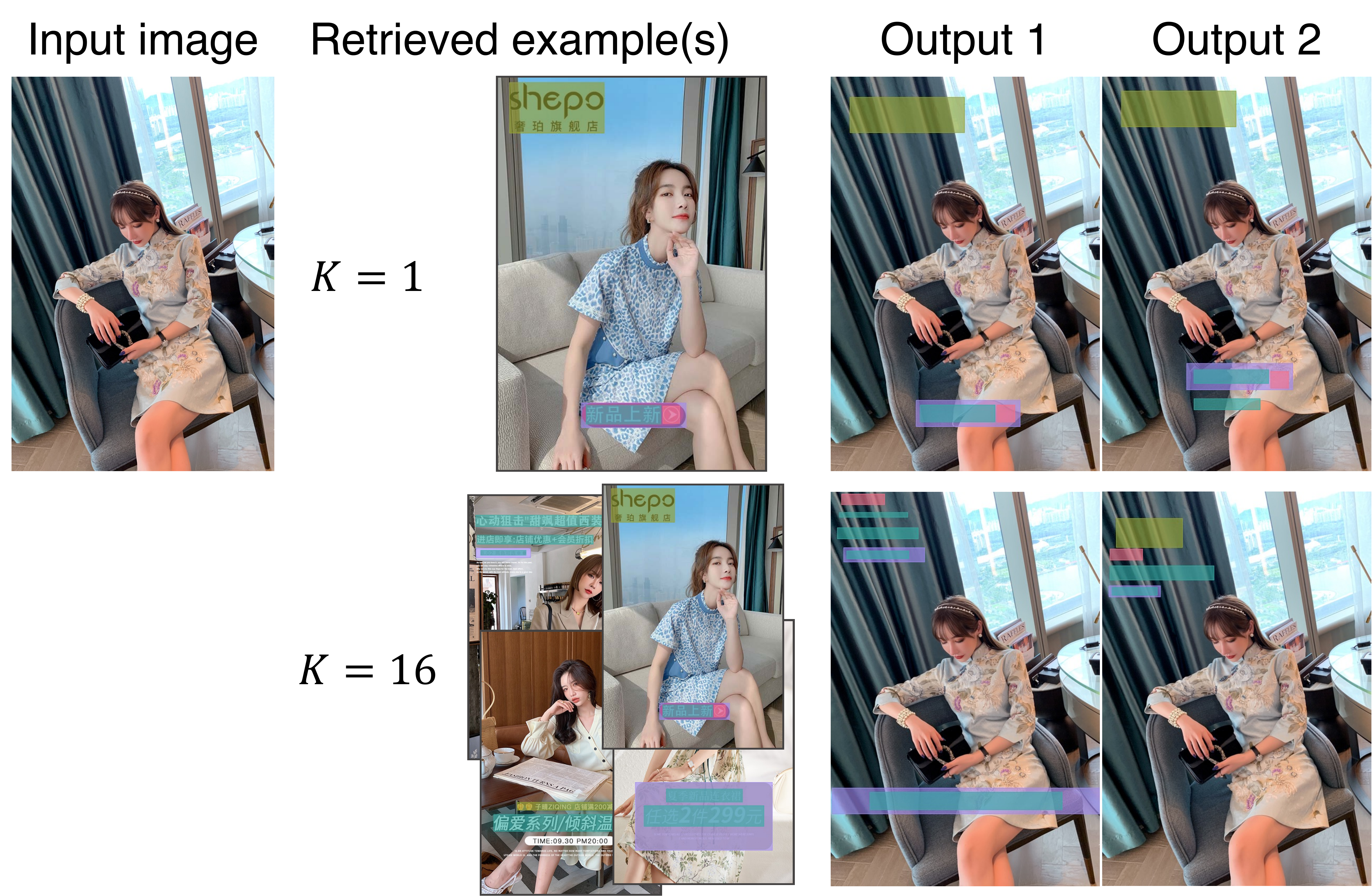}
    \caption{
        Visual comparison of retrieval and generated layouts with different retrieval sizes ($K\!=\!1$ and $16$).
        We display the top-4 examples for $K\!=\!16$ due to the limited space. 
        The output layouts are generated using different random seeds for variety.
    }
    \label{fig:retrieval_size_qualitative}
    \setlength\tabcolsep{6pt}
  \end{figure}
}

\newcommand{\tabFigMulti}{

    \begin{figure*}[t]
        \begin{minipage}{0.27\linewidth}
            \centering
            \includegraphics[width=0.8\linewidth]{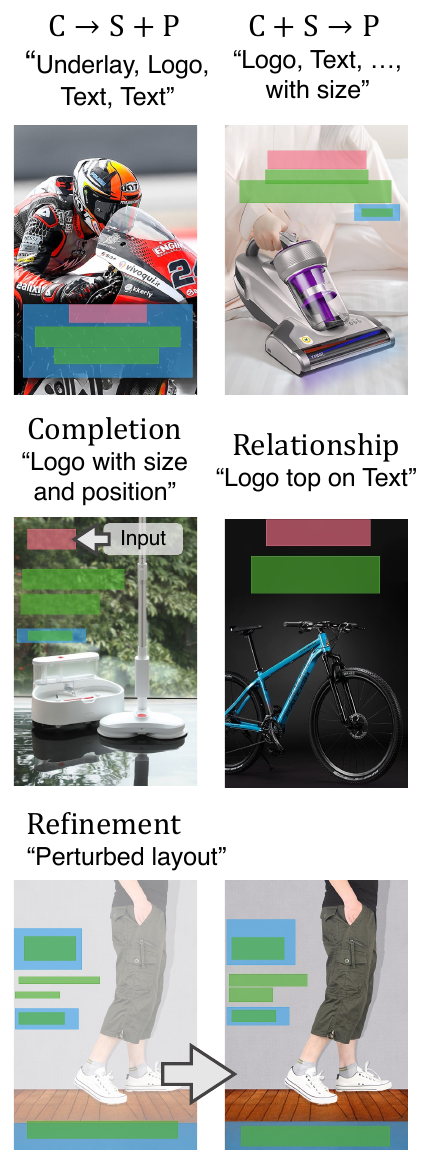}
            \captionof{figure}{
            Examples of input constraints and generated results for each constrained generation task. Quotation marks indicate the constraints.
            }
            \label{minifig:explain_constrained_task}
        \end{minipage}
        \hfill
        \begin{minipage}{0.71\linewidth}
            
            \resizebox{\linewidth}{!}{
                    \centering
                    \begin{tabular}{@{}l c c c c c c c c c c@{}}
                    \toprule
                    \multirow{3}{*}{Method}  
                & \multicolumn{5}{c}{PKU} & \multicolumn{5}{c}{CGL} \\
                \cmidrule(lr){2-6} \cmidrule(lr){7-11}
                ~ &
                \multicolumn{2}{c}{Content} &\multicolumn{3}{c}{Graphic} & 
                \multicolumn{2}{c}{Content} & \multicolumn{3}{c}{Graphic} \\
                \cmidrule(lr){2-3} \cmidrule(lr){4-6}
                \cmidrule(lr){7-8} \cmidrule(lr){9-11}
                ~ & 
                Occ $\downarrow$ & Rea $\downarrow$ & Und $\uparrow$ & Ove $\downarrow$ & FID$\downarrow$
                & Occ $\downarrow$ & Rea $\downarrow$ & Und $\uparrow$ & Ove $\downarrow$ & FID$\downarrow$ \\
                    \midrule
                
                    $\boldsymbol{\mathrm{C} \rightarrow \mathrm{S} + \mathrm{P} }$ \\
                
                    CGL-GAN
                    & \E{0.132} & \E{0.0158} & \E{0.48} & 0.038 & 11.47
                    & 0.140 & 0.0213 & 0.65 & 0.047  & 23.93
                    \\

                    LayoutDM$^{\dagger}$
                    & 0.152 & 0.0201 & 0.46 & 0.172  & 20.56
                    & 0.127 & 0.0192 & 0.79 & 0.026  & 3.39
                    \\
                
                    Autoreg Baseline
                    & 0.135 & 0.0167 & 0.43 & \E{0.028}  & \E{10.48}
                    & \D{0.124} & \E{0.0188} & \E{0.89} & \E{0.015}  & \E{1.36}
                    \\
                    \gb{RALF (Ours)}
                    & \gb{\D{0.124}} & \gb{\D{0.0138}} & \gb{\D{0.90}} & \gb{\D{0.010}}  & \gb{\D{2.21}}
                    & \gb{\E{0.126}} & \gb{\D{0.0180}} & \gb{\D{0.97}} & \gb{\D{0.006}}  & \gb{\D{0.50}}
                    \\
                    
                    \midrule

                    $\boldsymbol{ \mathrm{C} + \mathrm{S} \rightarrow \mathrm{P} }$ \\
                    
                    CGL-GAN
                    & \E{0.129} & \E{0.0155} & \E{0.48} & 0.043  & 9.11 
                    & \E{0.129} & 0.0202 & 0.75 & 0.027  & 6.96 
                    \\
                
                    LayoutDM$^{\dagger}$
                    & 0.143 & 0.0185 & 0.45 & 0.122  & 24.90
                    & \D{0.127} & \E{0.0190} & 0.82 & 0.021  & 2.18 
                    \\
                
                    Autoreg Baseline
                    & 0.137 & 0.0169 & 0.46 & \E{0.028} & \E{5.46}
                    & \D{0.127} & 0.0191 & \E{0.88} & \E{0.013} & \E{0.47}
                    \\
                    \gb{RALF (Ours)} 
                    & \gb{\D{0.125}} & \gb{\D{0.0138}} & \gb{\D{0.87}} & \gb{\D{0.010}} & \gb{\D{0.62}} 
                    & \gb{0.128} & \gb{\D{0.0185}} & \gb{\D{0.96}} & \gb{\D{0.006}} & \gb{\D{0.21}}
                
                    \\
                    
                    \midrule

                    \textbf{Completion} \\
                
                    CGL-GAN 
                    & 0.150 & 0.0174 & 0.43 & 0.061 & 25.67
                    & 0.174 & 0.0231 & 0.21 & 0.182 & 78.44 
                    \\
                    
                    LayoutDM$^{\dagger}$ 
                    & 0.135 & 0.0175 & 0.35 & 0.134 & 21.70
                    & 0.127 & \E{0.0192} & 0.76 & 0.020 & 3.19
                    \\
                
                    Autoreg Baseline
                    & \E{0.125} & \E{0.0161} & \E{0.42} & \E{0.023} & \E{5.96}
                    & \D{0.124} & \D{0.0185} & \E{0.91} & \E{0.011} & \E{2.33}
                    \\
                    \gb{RALF (Ours)}
                    & \gb{\D{0.120}} & \gb{\D{0.0140}} & \gb{\D{0.88}} & \gb{\D{0.012}} & \gb{\D{1.58}} 
                    & \gb{\E{0.126}} & \gb{\D{0.0185}} & \gb{\D{0.96}} & \gb{\D{0.005}} & \gb{\D{1.04}}
                    \\
                    
                    \midrule

                    \textbf{Refinement} \\
                    
                    CGL-GAN  
                    & 0.122 & 0.0141 & 0.39 & 0.090 & 6.40  
                    & \D{0.124} & \E{0.0182} & 0.86 & 0.024 & \E{1.20}
                    \\
                    LayoutDM$^{\dagger}$  
                    & \E{0.115} & \E{0.0121} & \E{0.57} & \E{0.008} & \E{2.86}
                    & 0.127 & 0.0188 & 0.75 & 0.018 & 1.98
                    \\
                
                    Autoreg Baseline
                    & 0.131 & 0.0171 & 0.41 & 0.026 & 5.89
                    & \E{0.126} & 0.0183 & \E{0.89} & \E{0.004} & 0.15
                    \\
                    \gb{RALF (Ours)}
                    & \gb{\D{0.113}} & \gb{\D{0.0109}} & \gb{\D{0.95}} & \gb{\D{0.004}} & \gb{\D{0.13}}
                    & \gb{\E{0.126}} & \gb{\D{0.0176}} & \gb{\D{0.98}} & \gb{\D{0.002}} & \gb{\D{0.14}}
                    \\
                        
                    \midrule
                
                    \textbf{Relationship} \\

                    Autoreg Baseline
                    & 0.140 & 0.0177 & 0.44 & 0.028 & 10.61
                    & 0.127 & 0.0189 & 0.88 & 0.015 & 1.28
                    \\
                    \gb{RALF (Ours)}
                    & \gb{\D{0.122}} & \gb{\D{0.0141}} & \gb{\D{0.85}} & \gb{\D{0.009}} & \gb{\D{2.23}}
                    & \gb{\D{0.126}} & \gb{\D{0.0184}} & \gb{\D{0.95}} & \gb{\D{0.006}} & \gb{\D{0.55}}
                    \\
                
                    \bottomrule
                    
                \end{tabular}
            }

            \captionof{table}{
            Quantitative result of six constrained generation tasks on the PKU and CGL test split.
            }
            \label{minitab:multitask}
        \end{minipage}
    \end{figure*}
}

\newcommand{\figCGLSpatialDiff}{
\begin{figure}
    \centering
    \includegraphics[width=\linewidth]{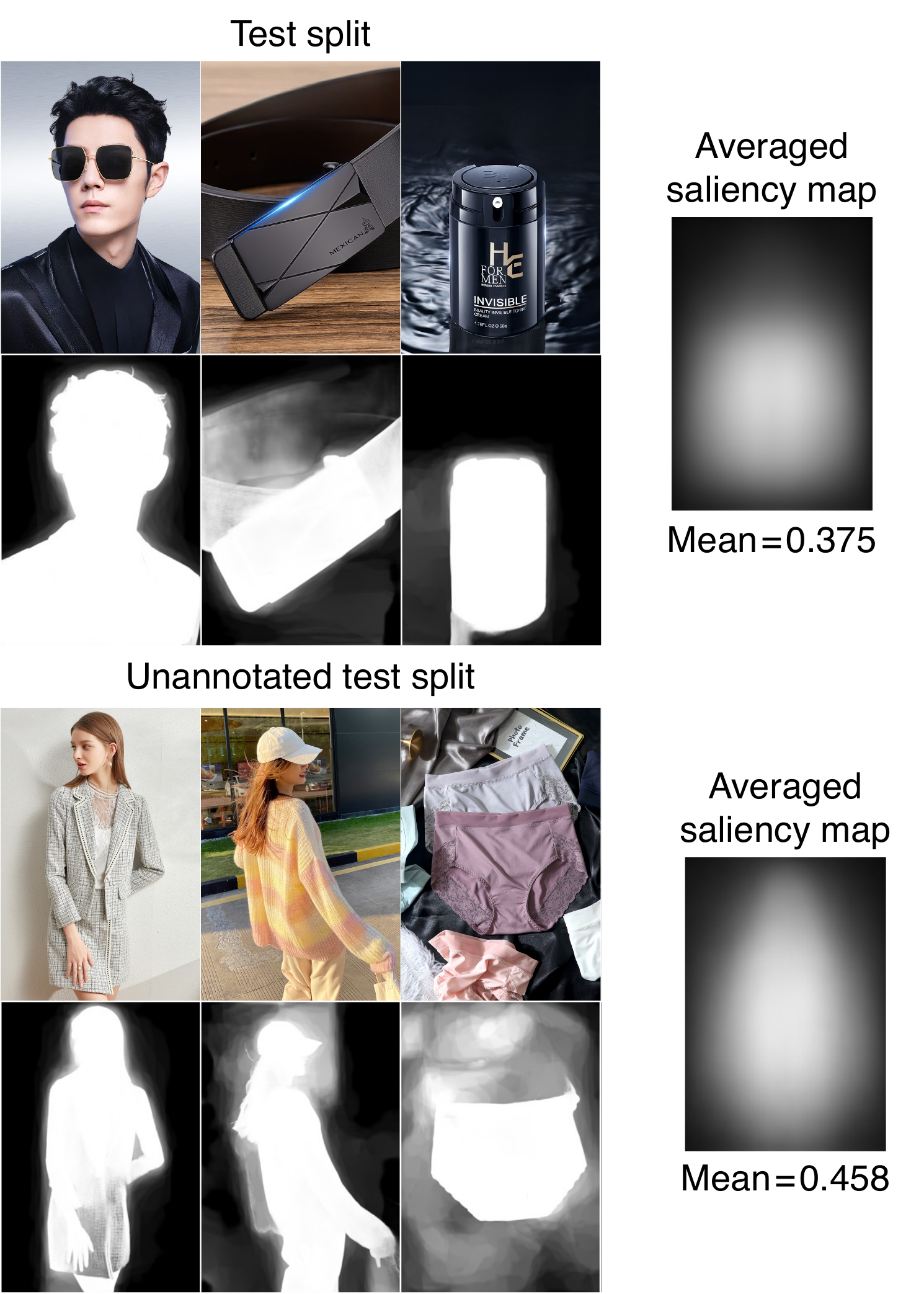}
    \caption{
        Visual comparison of canvases and saliency maps between the test and unannotated test split of the CGL dataset.
        Canvases are randomly selected from each split.
        The averaged saliency map is produced by computing the spatial average of all saliency maps of each split.
        Mean represents the spatial average of all saliency maps of each split.
    }
    \label{fig:cgl_spatial_diff}
  \end{figure}
}

\newcommand{\tabChangeParameter}{
  \begin{table*}[t]
    \centering
\resizebox{\linewidth}{!}{

\begin{tabular}{@{}l c c c c c c c c c c c c@{}}
\toprule
\multirow{3}{*}{Method} & \multirow{3}{*}{\#Dim} & \multirow{3}{*}{\#ParamsDec}  & \multicolumn{5}{c}{PKU} & \multicolumn{5}{c}{CGL} \\
\cmidrule(lr){4-8} \cmidrule(lr){9-13}
~ & ~ &~ &
\multicolumn{2}{c}{Content} &\multicolumn{3}{c}{Graphic} & 
\multicolumn{2}{c}{Content} & \multicolumn{3}{c}{Graphic} \\
\cmidrule(lr){4-5} \cmidrule(lr){6-8}
\cmidrule(lr){9-10} \cmidrule(lr){11-13}
~ & ~ & ~ 
& Occ $\downarrow$ & Rea $\downarrow$ & Und $\uparrow$ & Ove $\downarrow$ & FID$\downarrow$
& Occ $\downarrow$ & Rea $\downarrow$ & Und $\uparrow$ & Ove $\downarrow$ & FID$\downarrow$ \\
\midrule

Autoreg Baseline & \multirow{2}{*}{128} & \multirow{2}{*}{2.55M}
 & 0.146 & 0.0184 & 0.41 & 0.030 & 18.86 
 & 0.127 & 0.0196 & 0.86 & 0.013 & 3.60 
\\

RALF & ~ & ~
 & 0.123 & 0.0141 & 0.71 & 0.007 & 4.14
 & 0.125 & 0.0180 & 0.97 & 0.005 & 1.27
\\

\midrule

Autoreg Baseline$^{\diamondsuit}$ & \multirow{2}{*}{256} & \multirow{2}{*}{6.59M}
 & \E{0.134} & \E{0.0165} & 0.44 & \E{0.018} & 13.51
 & 0.125 & \E{0.0190} & \E{0.92} & \E{0.011} & 2.90
\\

\gb{RALF$^{\diamondsuit}$} & ~ & ~
 & \gb{0.119} & \gb{0.0129} & \gb{0.92} & \gb{0.008} & \gb{3.45} 
 & \gb{0.125} & \gb{0.0180} & \gb{0.98} & \gb{0.004} & \gb{1.31}
\\

\midrule

Autoreg Baseline & \multirow{2}{*}{512} & \multirow{2}{*}{19.46M}
 & 0.128 & 0.0150 & 0.57 & 0.011 & 10.85
 & 0.122 & 0.0184 & 0.95 & 0.009 & 2.74  
\\

RALF & ~ & ~
 & 0.122 & 0.0131 & 0.94 & 0.010 & 3.61 
 &  0.128 & 0.0182 & 0.97 & 0.004 & 1.72
\\

\midrule

Autoreg Baseline & \multirow{2}{*}{768} & \multirow{2}{*}{38.82M}
 & 0.122 & 0.0150 & 0.70 & 0.012 & 8.46
 & 0.124 & 0.0183 & 0.95 & 0.008 & 2.26
\\

RALF & ~ & ~
 & 0.126 & 0.0131 & 0.93 & 0.008 & 3.19
 &  0.131 & 0.0187 & 0.97 & 0.004 & 1.72 
\\

\bottomrule
				\end{tabular}
    }
    \caption{
      Qualitative result of varying network parameters on unconstrained generation metrics on the PKU and CGL test split.
      We modify the number of features (\#Dim) in the input of cross-attention layers and the sequence to the decoder layer.
      \#ParamsDec indicates the number of parameters of the layout decoder.
      $\diamondsuit$ represents the setting of our experiments in the main manuscript.
    }
    \label{tab:ablation_dec_parameters}
  \end{table*}
}

\newcommand{\figInpaintingMain}{
\begin{figure}
    \centering
    \includegraphics[width=\linewidth]{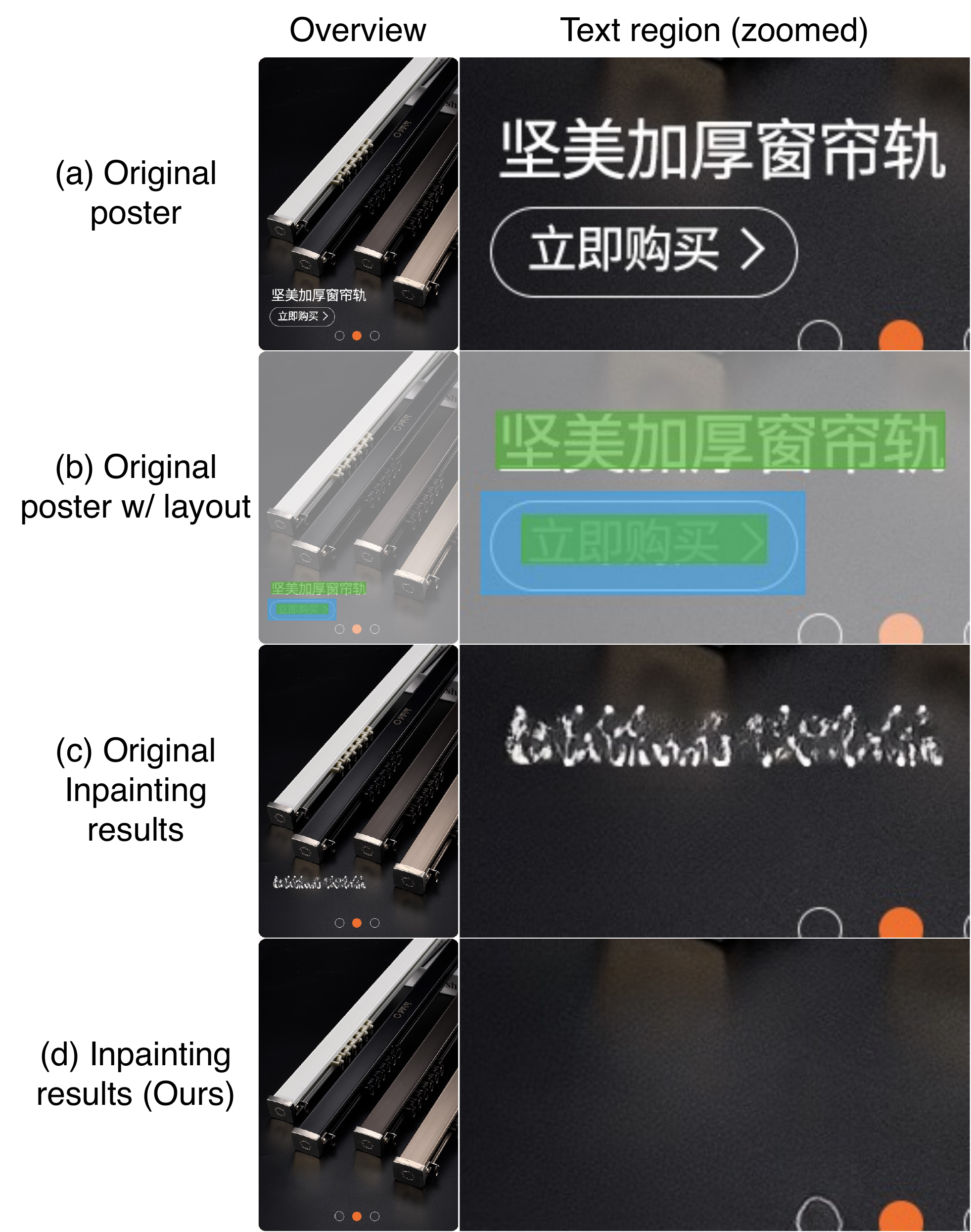}
    \caption{
        Comparison of inpainting for the dataset preprocessing.
    }
    \label{fig:better_preprocessing}
  \end{figure}
}

\newcommand{\figInpaintingSupps}{
\begin{figure}
    \centering
    \includegraphics[width=\linewidth]{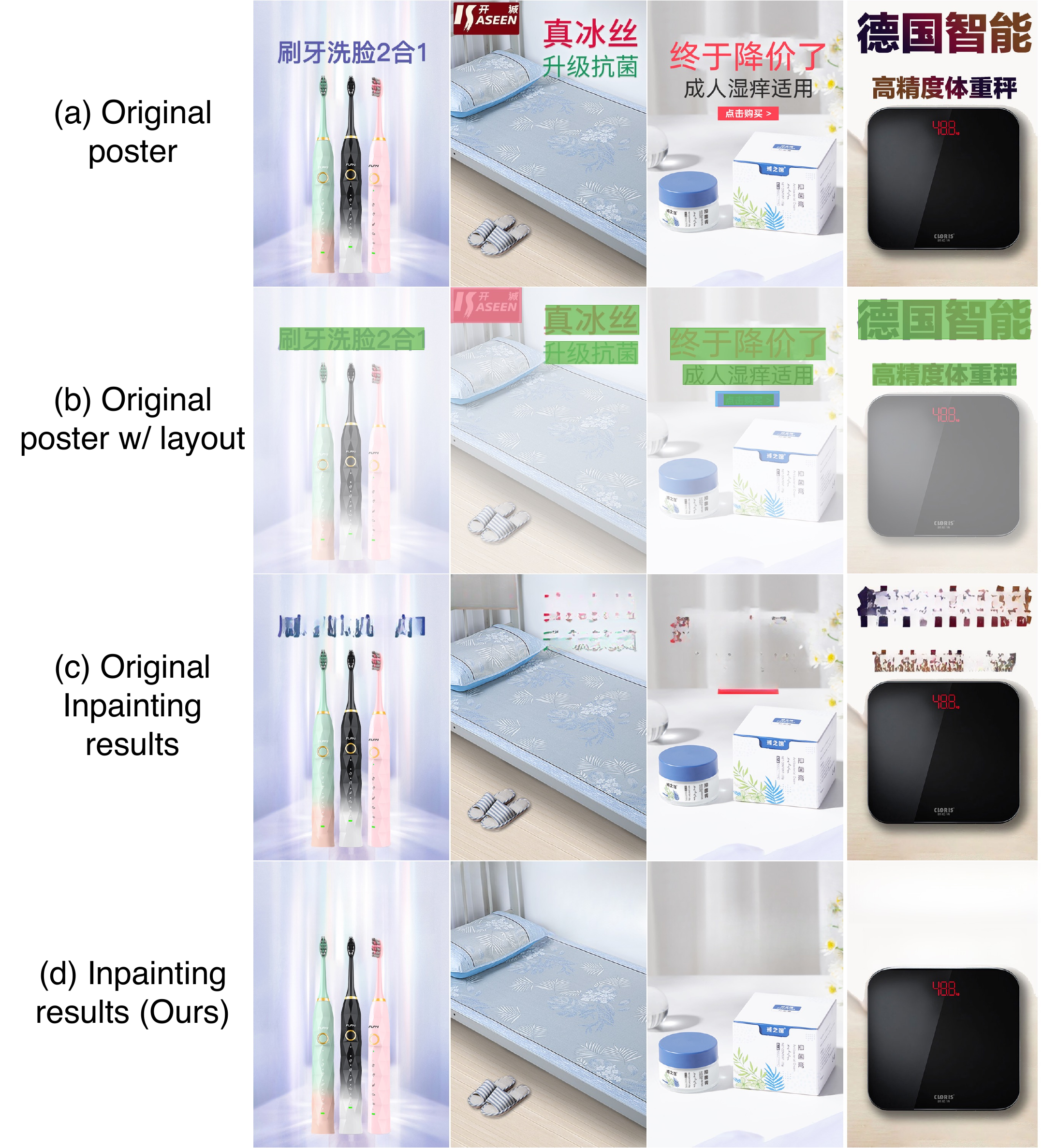}
    \vspace {-5.4mm}
    \caption{
        Comparison of inpainting for the dataset preprocessing.
    }
    \label{fig:better_preprocessing_2}
  \end{figure}
}

\newcommand{\figPKUNumElements}{
\begin{figure}
    \includegraphics[width=0.8\hsize]{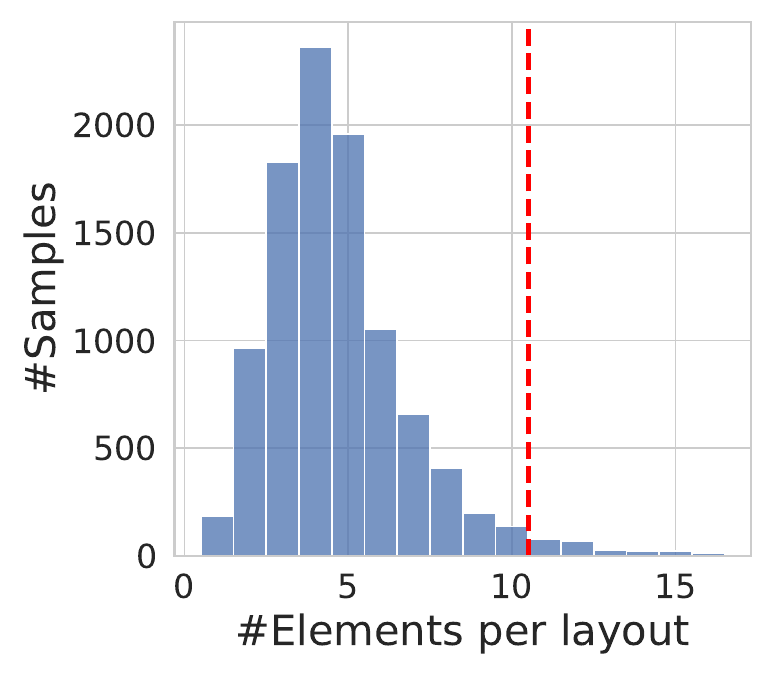}
    \vspace{-3mm}
    \caption{Number of elements per layout in the original PKU dataset. A red dashed line indicates the maximum number of elements we use.}
    \label{fig:num_elements_pku}
\end{figure}
}

\newcommand{\figVisualComparisonConstraintPKU}{
\begin{figure*}[t]
    \includegraphics[width=\hsize]{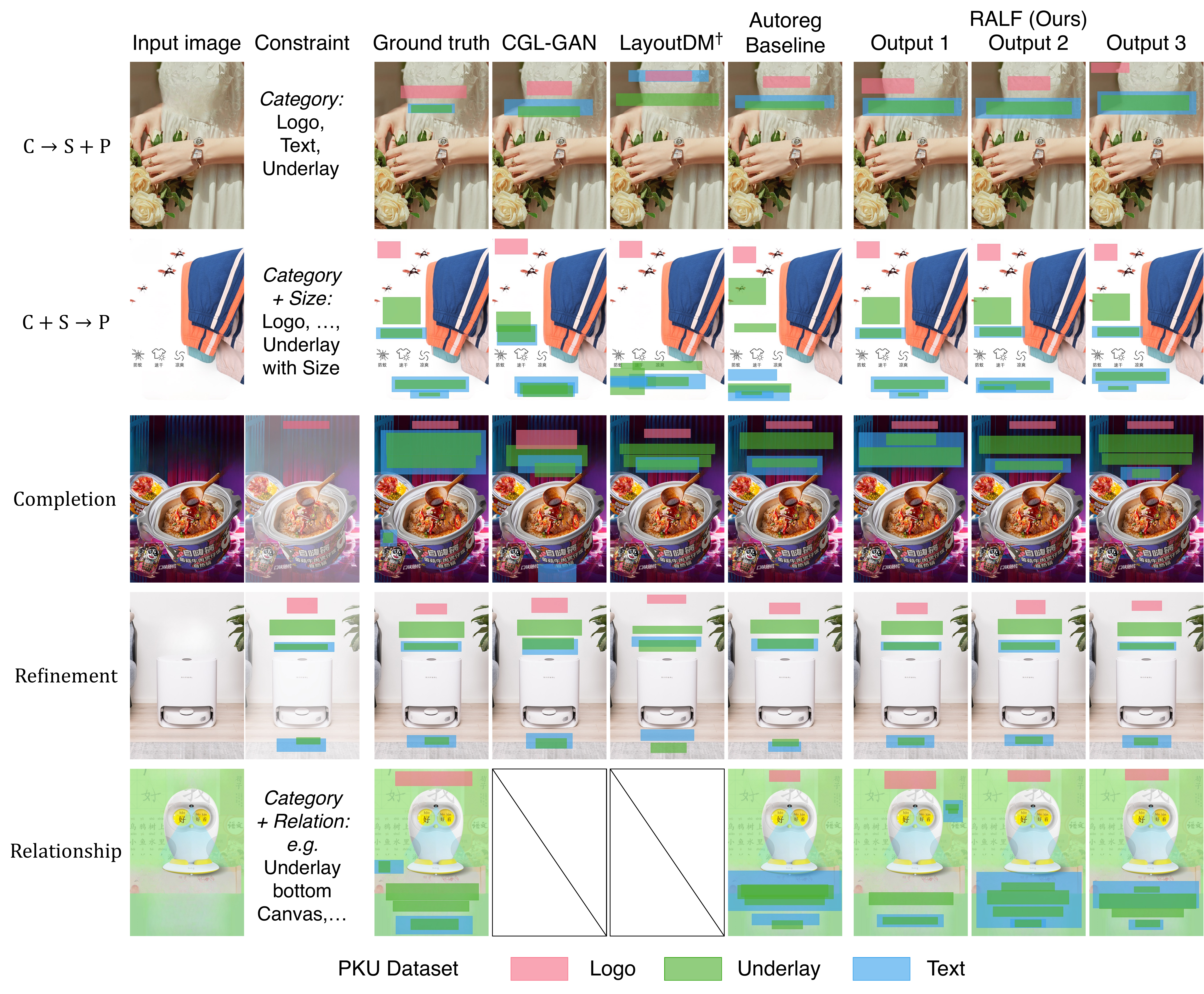}
    \caption{
     Visual comparison of constrained generation with
baselines on the PKU annotated test split.
    }
    \label{fig:visual_comparison_constraint_pku}
\end{figure*}
}

\newcommand{\figVisualComparisonConstraintCGL}{
\begin{figure*}[t]
    \includegraphics[width=\hsize]{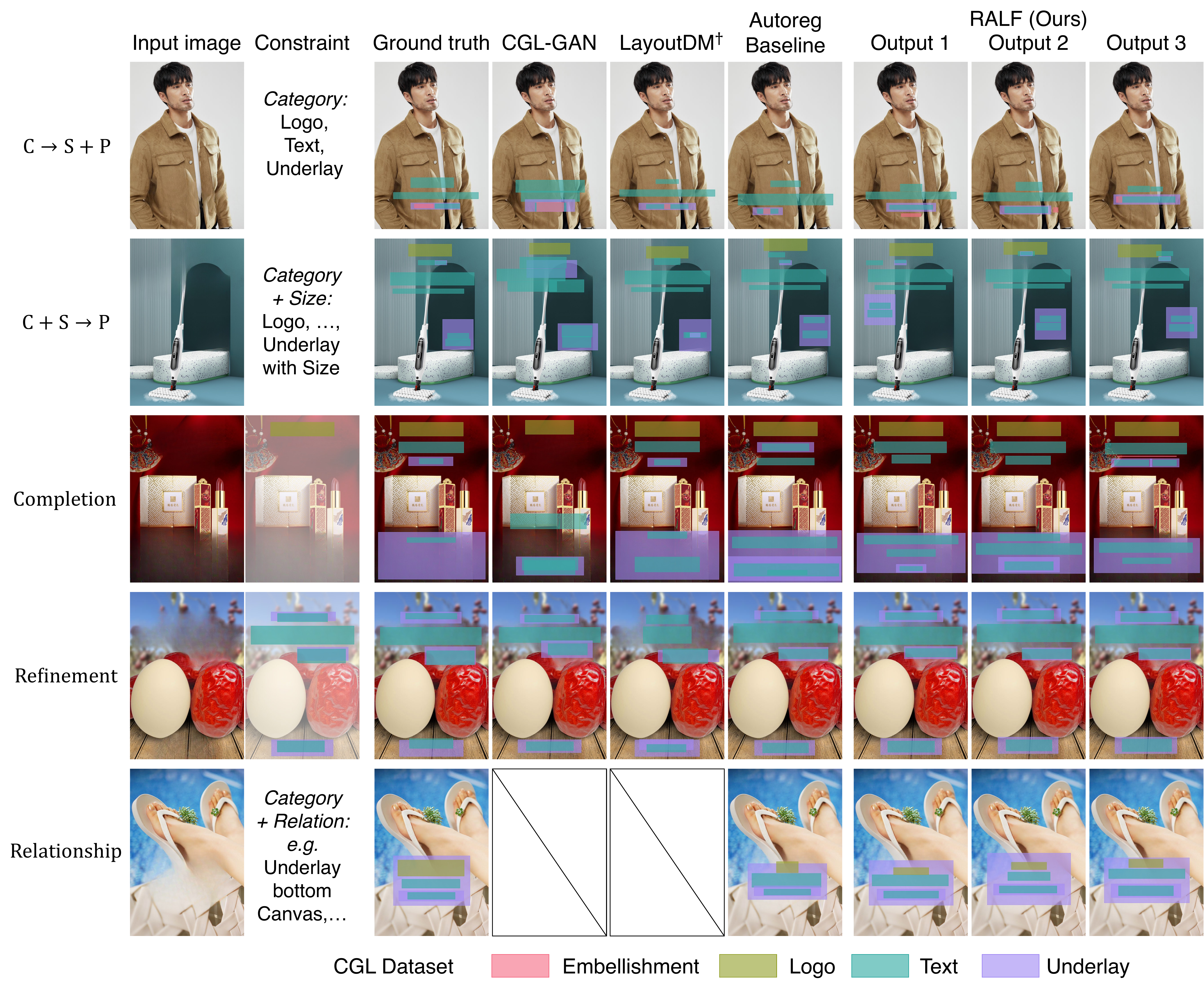}
    \caption{Visual comparison of constrained generation with
baselines on the CGL annotated test split.}
    \label{fig:visual_comparison_constraint_cgl}
\end{figure*}
}

\newcommand{\tabDetailedUncondPKU}{
  \begin{table*}[t]
    \centering
    \resizebox{0.9\linewidth}{!}{
    
    \begin{tabular}{@{}l c c c c c c c c c c@{}}
    \toprule
    \multirow{3}{*}{Method}  & \multicolumn{10}{c}{PKU} \\
\cmidrule(lr){2-11}
~ & \multicolumn{3}{c}{Content} & \multicolumn{7}{c}{Graphic} \\
\cmidrule(lr){2-4} \cmidrule(lr){5-11}
~ & Occ $\downarrow$ & Rea $\downarrow$ & R$_{\mathrm{shm}}$ $\downarrow$ & Align $\downarrow$ & Und$_{\mathrm{L}}$ $\uparrow$ & Und$_{\mathrm{S}}$ $\uparrow$ & Ove $\downarrow$ & Den$\uparrow$& Cov$\uparrow$ & FID$\downarrow$ \\
    \midrule

\tp{Real-Data} & \tp{0.112} & \tp{0.0102} & \tp{13.94} & \tp{0.00379} & \tp{0.99} & \tp{0.99} & \tp{0.0009} & \tp{0.95} & \tp{0.95} & \tp{1.58} \\
\tp{Top1-Retrieval} & \tp{0.212} & \tp{0.0218} & \tp{16.33} & \tp{0.00371} & \tp{0.99} & \tp{0.99} & \tp{0.002} & \tp{1.07} & \tp{0.97} & \tp{1.43} \\
CGL-GAN~\cite{cglgan} & 0.138 & 0.0164 & 14.32 & 0.00311 & 0.81 & 0.41 & 0.074 & 0.70 & 0.68 & 34.51 \\
DS-GAN~\cite{pkuposter} & 0.142 & 0.0169 & 14.95 & 0.00347 & 0.89 & 0.63 & 0.027 & 1.10 & 0.82 & 11.80 \\
ICVT~\cite{cao2022geometry} & 0.146 & 0.0185 & 13.92 & 0.00228 & 0.63 & 0.49 & 0.318 & 0.35 & 0.40 & 39.13 \\
LayoutDM$^{\dagger}$~\cite{inoue2023layout} & 0.150 & 0.0192 & \D{13.06} & 0.00298 & 0.64 & 0.41 & 0.190 & 0.74 & 0.59 & 27.09 \\
Autoreg Baseline & 0.134 & 0.0164 & 14.43 & \D{0.00192} & 0.79 & 0.43 & 0.019 & 1.13 & 0.79 & 13.59 \\
\gb{RALF (Ours)} & \gb{\D{0.119}} & \gb{\D{0.0129}} & \gb{14.11} & \gb{0.00267} & \gb{\D{0.98}} & \gb{\D{0.92}} & \gb{\D{0.008}} & \gb{\D{1.25}} & \gb{\D{0.97}} & \gb{\D{3.45}} \\

    \bottomrule
    
    				\end{tabular}
        }
        \caption{
          Unconstrained generation results on the PKU test split.
        }
        \label{tab:detailed_uncond_pku}
  \end{table*}
}

\newcommand{\tabDetailedUncondCGL}{
  \begin{table*}[t]
    \centering
    \resizebox{0.9\linewidth}{!}{
    
    \begin{tabular}{@{}l c c c c c c c c c c@{}}
    \toprule
    \multirow{3}{*}{Method}  & \multicolumn{10}{c}{CGL} \\
\cmidrule(lr){2-11}
~ & \multicolumn{3}{c}{Content} & \multicolumn{7}{c}{Graphic} \\
\cmidrule(lr){2-4} \cmidrule(lr){5-11}
~ & Occ $\downarrow$ & Rea $\downarrow$ & R$_{\mathrm{shm}}$ $\downarrow$ & Align $\downarrow$ & Und$_{\mathrm{L}}$ $\uparrow$ & Und$_{\mathrm{S}}$ $\uparrow$ & Ove $\downarrow$ & Den$\uparrow$& Cov$\uparrow$ & FID$\downarrow$ \\
    \midrule

\tp{Real-Data} & \tp{0.125} & \tp{0.0170} & \tp{14.33} & \tp{0.00240} & \tp{0.99} & \tp{0.98} & \tp{0.0002} & \tp{0.93} & \tp{1.00} & \tp{0.79} \\

\tp{Top1-Retrieval} & \tp{0.214} & \tp{0.0266} & \tp{16.02} & \tp{0.00254} & \tp{0.99} & \tp{0.99} & \tp{0.0005} & \tp{1.01} & \tp{0.90} & \tp{0.93} \\

CGL-GAN~\cite{cglgan} & 0.157 & 0.0237 & 14.12 & 0.00320 & 0.67 & 0.29 & 0.161 & 0.31 & 0.28 & 66.75 \\

DS-GAN~\cite{pkuposter} & 0.141 & 0.0229 & 14.85 & 0.00257 & 0.71 & 0.45 & 0.057 & 0.64 & 0.40 & 41.57 \\

ICVT~\cite{cao2022geometry} & \D{0.124} & 0.0205 & \D{13.40} & 0.00319 & 0.55 & 0.42 & 0.310 & 0.16 & 0.22 & 65.34 \\

LayoutDM$^{\dagger}$~\cite{inoue2023layout} & 0.127 & 0.0192 & 14.15 & 0.00242 & 0.92 & 0.82 & 0.020 & 0.87 & 0.93 & 2.36 \\

Autoreg Baseline & 0.125 & 0.0190 & 14.22 & \D{0.00234} & 0.97 & 0.92 & 0.011 & 1.05 & 0.91 & 2.89 \\

\gb{RALF (Ours)} & \gb{0.125} & \gb{\D{0.0180}} & \gb{14.26} & \gb{0.00236} & \gb{\D{0.99}} & \gb{\D{0.98}} & \gb{\D{0.004}} & \gb{\D{1.09}} & \gb{\D{0.96}} & \gb{\D{1.32}} \\
    \bottomrule
    
    				\end{tabular}
        }
        \caption{
          Unconstrained generation results on the CGL test split.
        }
        \label{tab:detailed_uncond_cgl}
  \end{table*}
}

\newcommand{\tabmultitaskwithRAClear}{
  \begin{table*}[t]
    \centering
    \resizebox{\linewidth}{!}{
    
    \begin{tabular}{@{}l c c c c c c c c c c c c@{}}
    \toprule
    \multirow{3}{*}{Task} & \multirow{3}{*}{Method} & \multirow{3}{*}{Retrieval}  & \multicolumn{5}{c}{PKU} & \multicolumn{5}{c}{CGL} \\
\cmidrule(lr){4-8} \cmidrule(lr){9-13}
~ & ~ & ~ & \multicolumn{2}{c}{Content} & \multicolumn{3}{c}{Graphic} & \multicolumn{2}{c}{Content} & \multicolumn{3}{c}{Graphic} \\
\cmidrule(lr){4-5} \cmidrule(lr){6-8}
\cmidrule(lr){9-10} \cmidrule(lr){11-13}
~ & ~ & ~ & Occ $\downarrow$ & Rea $\downarrow$ & Und $\uparrow$ & Ove $\downarrow$ & FID$\downarrow$
          & Occ $\downarrow$ & Rea $\downarrow$ & Und $\uparrow$ & Ove $\downarrow$ & FID$\downarrow$ \\
    \midrule

    \multirow{6}{*}{Unconstraint}

    ~ & \tp{Real Data} & ~
    & \tp{0.112} & \tp{0.0102} & \tp{0.99} & \tp{0.0009} & \tp{1.58} 
    & \tp{0.125} & \tp{0.0170} & \tp{0.98} & \tp{0.0002} & \tp{0.79}
    \\
    
    ~ & \tp{Top-1 Retrieval} & ~
    & \tp{0.212} & \tp{0.0218} & \tp{0.99} & \tp{0.002} & \tp{1.43} 
    & \tp{0.214} & \tp{0.0266} & \tp{0.99} & \tp{0.0005} & \tp{0.93}
    \\
    
    ~ & CGL-GAN & ~
    & 0.138 & 0.0164 & 0.41 & 0.074 & 34.51
    & 0.157 & 0.0237 & 0.29 & 0.161 & 66.75
    \\
    
    ~ & \F{CGL-GAN} & \F{\ding{51}}
    & \F{0.144} & \F{0.0164} & \F{0.63} & \F{0.039} & \F{13.28}
    & \F{0.172} & \F{0.0245} & \F{0.42} & \F{0.157} & \F{60.67}
    \\
    
    ~ & LayoutDM$^{\dagger}$ & ~
    & 0.150 & 0.0192 & 0.41 & 0.190 & 27.09
    & 0.127 & 0.0192 & 0.82 & 0.020 & 2.36
    \\
    
    ~ & \F{LayoutDM$^{\dagger}$} & \F{\ding{51}}
    & \F{0.123} & \F{0.0144} & \F{0.51} & \F{0.091} & \F{10.03}
    & \F{0.126} & \F{0.0187} & \F{0.85} & \F{0.019} & \F{1.97}
    \\
    
    \midrule

    \multirow{4}{*}{$\mathrm{C} \rightarrow \mathrm{S} + \mathrm{P} $}
    
    ~ & CGL-GAN & ~ 
    & 0.132 & 0.0158 & 0.48 & 0.038 & 11.47
    & 0.140 & 0.0213 & 0.65 & 0.047 & 23.93
    \\
    ~ & \F{CGL-GAN} & \F{\ding{51}}
    & \F{0.140} & \F{0.0153} & \F{0.66} & \F{0.030} & \F{10.23}
    & \F{0.138} & \F{0.0202} & \F{0.82} & \F{0.021} & \F{10.01}
    \\
    ~ & LayoutDM$^{\dagger}$ & ~
    & 0.152 & 0.0201 & 0.46 & 0.172 & 20.50
    & 0.127 & 0.0192 & 0.79 & 0.026 & 3.39
    \\
    ~ & \F{LayoutDM$^{\dagger}$} & \F{\ding{51}}
    & \F{0.121} & \F{0.0141} & \F{0.55} & \F{0.088} & \F{9.02}
    & \F{0.127} & \F{0.0189} & \F{0.81} & \F{0.026} & \F{3.36 }
    \\ 
    
    \midrule

    \multirow{4}{*}{$\mathrm{C} + \mathrm{S} \rightarrow \mathrm{P} $}
    
    & CGL-GAN & ~
    & 0.129 & 0.0155 & 0.48 & 0.043 & 9.11
    & 0.129 & 0.0202 & 0.75 & 0.027 & 6.96
    \\
    & \F{CGL-GAN} & \F{\ding{51}}
    & \F{0.146} & \F{0.0178} & \F{0.57} & \F{0.036} & \F{7.74}
    & \F{0.135} & \F{0.0207} & \F{0.78} & \F{0.020} & \F{6.01}
    \\
    
    & LayoutDM$^{\dagger}$  & ~
    & 0.143 & 0.0185 & 0.45 & 0.122 & 24.90
    & 0.127 & 0.0190 & 0.82 & 0.021 & 2.18
    \\

    & \F{LayoutDM$^{\dagger}$} & \F{\ding{51}}
    & \F{0.123} & \F{0.0144} & \F{0.59} & \F{0.071} & \F{10.68 }
    & \F{0.127} & \F{0.0188} & \F{0.83} & \F{0.020} & \F{1.77 }
    \\
    
    \midrule

    \multirow{4}{*}{Completion}
    
    & CGL-GAN & ~
    & 0.146 & 0.0175 & 0.42 & 0.076  & 27.18
    & 0.174 & 0.0231 & 0.21 & 0.182 & 78.44
    \\

    & \F{CGL-GAN} & \F{\ding{51}}
    & \F{0.146} & \F{0.0169} & \F{0.71} & \F{0.039} & \F{12.46}
    & \F{0.155} & \F{0.0230} & \F{0.46} & \F{0.102} & \F{48.82}
    \\
    
    & LayoutDM$^{\dagger}$    & ~  
    & 0.135 & 0.0175 & 0.35 & 0.134 & 21.70
    & 0.127 & 0.0192 & 0.76 & 0.020 & 3.19
    \\

    & \F{LayoutDM$^{\dagger}$} & \F{\ding{51}}
    & \F{0.120} & \F{0.0143} & \F{0.45} & \F{0.071} & \F{12.96}  
    & \F{0.126} & \F{0.0189} & \F{0.79} & \F{0.018} & \F{2.55}
    \\

    \midrule

    \multirow{4}{*}{Refinement}
    
    & CGL-GAN & ~
    & 0.122 & 0.0141 & 0.39 & 0.090 & 6.40
    & 0.124 & 0.0182 & 0.86 & 0.024 & 1.20
    \\

    & \F{CGL-GAN} & \F{\ding{51}}
    & \F{0.129} & \F{0.0157} & \F{0.37} & \F{0.072} & \F{4.91}
    & \F{0.133} & \F{0.0194} & \F{0.85} & \F{0.013} & \F{1.56}
    \\
    
    & LayoutDM$^{\dagger}$  & ~
    & 0.115 & 0.0121 & 0.57 & 0.008 & 2.86
    & 0.127 & 0.0188 & 0.75 & 0.018 & 1.98
    \\

    & \F{LayoutDM$^{\dagger}$} & \F{\ding{51}}
    & \F{0.115} & \F{0.0121} & \F{0.57} & \F{0.007} & \F{2.91 }
    & \F{0.126} & \F{0.0186} & \F{0.76} & \F{0.019} & \F{1.79 }
    \\

    \bottomrule
    
    				\end{tabular}
        }
        \caption{
          Retrieval augmentation for CGL-GAN and LayoutDM$^{\dagger}$ on the PKU and CGL test split for unconstrained and constrained generation.
        }
        \label{tab:multitask_for_previous_methods}
  \end{table*}
}

\newcommand{\tabComparingInferenceSpeed}{
  \begin{table}[t]
    \centering
\resizebox{1.0\linewidth}{!}{

\begin{tabular}{@{}l c c c c c c c c c c @{}}
\toprule
~ & \multirow{2}{*}{CGL-GAN} & \multirow{2}{*}{LayoutDM$^{\dagger}$}
 & \multirow{2}{*}{\parbox{1.2cm}{Autoreg \\ Baseline}}  & \multicolumn{3}{c}{RALF}
\\
\cmidrule(lr){5-8}
~ & ~ & ~ & ~ & DreamSim & Retrieval & Network & \multicolumn{1}{|c}{Total} \\
\midrule
Time~[s] & 0.012 & 0.495 & 0.225 & 0.022 & 0.031 & 0.252 &   \multicolumn{3}{|c}{0.305} \\ 

\bottomrule
				\end{tabular}
    }
    \vspace{-0.2cm}
    \caption{
    Inference time comparison on the PKU dataset.
      RALF consists of three components -- feature extraction (DreamSim), layout retrieval (Retrieval), and layout generation (Network).
    The total inference time (Total) is the sum of these individual components.
    }
    \label{tab:comparing_inference_speed}
  \end{table}
}

\newcommand{\tabInaccurateSaliencyMap}{
  \begin{table}[t]
    \centering
\resizebox{1.0\linewidth}{!}{

\begin{tabular}{@{}l c c c c c c @{}}
\toprule
Method & Saliency map & Occ $\downarrow$ & Rea $\downarrow$ & Und $\uparrow$ & Ove $\downarrow$ & FID $\downarrow$ \\

\midrule
Autoreg Baseline &  ~        & 0.132 & 0.0169 & 0.45 & 0.021 & 11.78 \\
Autoreg Baseline & \ding{51} & 0.134 & 0.0165 & 0.44 & 0.018 & 13.51 \\
\midrule
RALF & ~                     & 0.122 & 0.0129 & 0.90 & 0.007 & 3.97  \\
RALF & \ding{51}             & 0.119 & 0.0129 & 0.92 & 0.008 & 3.45 \\

\bottomrule
				\end{tabular}
    }
    \vspace{-0.2cm}
    \caption{
         Quantitative results without and with a saliency map.
    }
    \label{tab:inaccurate_saliency_map}
  \end{table}
}

\begin{document}
\maketitle

\begin{abstract}
   Content-aware graphic layout generation aims to automatically arrange visual elements along with a given content, such as an e-commerce product image.
In this paper, we argue that the current layout generation approaches suffer from the limited training data for the high-dimensional layout structure.
We show that a simple retrieval augmentation can significantly improve the generation quality.
Our model, which is named Retrieval-Augmented Layout Transformer (RALF), retrieves nearest neighbor layout examples based on an input image and feeds these results into an autoregressive generator.
Our model can apply retrieval augmentation to various controllable generation tasks and yield high-quality layouts within a unified architecture.
Our extensive experiments show that RALF successfully generates content-aware layouts in both constrained and unconstrained settings and significantly outperforms the baselines.\footnote{Our project page is available at \href{https://udonda.github.io/RALF/}{https://udonda.github.io/RALF/}}
\end{abstract}

\def\ours{OURS}

\section{Introduction}

Layout is an essential part of graphic design, where the aesthetic appeal relies on the harmonious arrangement and selection of visual elements such as logos and texts.
In real-world creative workflows, such as posters~\cite{vinci, designscape} and magazines~\cite{yang2016automatic, jahanian13iui} creation, designers typically work on a given subject; for example, creating an advertising poster of a specific product.
We call layout generation under such conditions \emph{content-aware} layout generation, where the goal is to generate diverse yet plausible arrangements of element bounding boxes that harmonize with the given background image (canvas).
Recent studies~\cite{zheng19tog, cglgan} show that generative models can produce content-aware layouts that respect aesthetic principles, such as avoiding overlaps~\cite{vinci}.
However, generated layouts often still suffer from artifacts, including misaligned underlay embellishment and text elements.
We hypothesize that current approaches based solely on generative models do not scale due to the scarcity of highly structured layout data.
Unlike public images on the Web, curating a large dataset of layered graphic designs is not a viable solution since designers typically create their work in proprietary authoring tools, such as Adobe Illustrator~\cite{illustrator}.

\teaser

Inspired by the fact that designers often refer to existing designs~\cite{getting_inspired},
we propose a retrieval-augmented generation method to address the challenges in the layout domain.
Recent literature shows that retrieval augmentation helps in enhancing the generation quality of language models~\cite{retro, realm} and image synthesis~\cite{ret_aug_ddpm, sheynin2022knndiffusion}, thanks to the ability to reference real examples in the limited data domain.
We argue that retrieval augmentation plays an important role in mitigating the data scarcity problem in content-aware layout generation.

We build \textbf{R}etrieval-\textbf{A}ugmented \textbf{L}ayout Trans\textbf{F}ormer (RALF), which is an autoregressive generator capable of referencing external layout examples.
RALF retrieves reference layouts by nearest neighbor search based on the appearance of the input and supplements the generation process (\cref{fig:teaser}).
Since the input canvas and retrieved layouts have different modalities, we use the cross-attention mechanism to augment the feature input to the generator.
Although we build RALF with an autoregressive approach, retrieval augmentation is also effective in other generation approaches such as diffusion models~\cite{inoue2023layout}, which we show in the experiments.

We evaluate our RALF on public benchmarks~\cite{pkuposter,cglgan} and show that RALF outperforms state-of-the-art models in content-aware layout generation.
Thanks to the retrieval capability,
RALF requires less than half the training data to achieve the same performance as the baseline.
We further show that our modular architecture can adapt to \emph{controllable} generation tasks that impose various user-specified constraints, which is common in real-world workflow.

\figFramework

We summarize our contributions as follows:
1) We find that retrieval augmentation effectively addresses the data scarcity problem in content-aware layout generation.
2) We propose a Retrieval-Augmented Layout Transformer (RALF) designed to integrate retrieval augmentation for layout generation tasks.
3) Our extensive evaluations show that our RALF successfully generates high-quality layouts under various scenarios and significantly outperforms baselines.
We will make our code publicly available on acceptance.

\section{Related Work}

\subsection{Content-agnostic Layout Generation}
Content-agnostic layout generation, which aims at generating layouts without a specific input canvas, has been studied for a long time~\cite{lok2001survey, agrawala2011design, designscape, yang2016automatic}.
The typical approach involves predicting the arrangement of elements, where each element has a tuple of attributes such as category, position, and size~\cite{li2019layoutgan}.
Recent approaches employ various types of neural networks-based generative models, such as generative adversarial networks (GAN)~\cite{li2019layoutgan,jianan2021tvcg,Kikuchi2021}, variational autoencoders (VAE)~\cite{jyothi2019layoutvae,arroyo2021variational,jiang2022coarse}, autoregressive models~\cite{gupta2021layouttransformer,layoutFormerPP}, non-autoregressive models~\cite{kong2021blt}, and diffusion models~\cite{inoue2023layout,chai2023layoutdm,zhang2023layoutdiffusion,levi2023dlt}.
Note that the retrieval augmentation discussed in this paper may not be directly applicable to the content-agnostic setup due to the lack of input queries.

Several works consider user-specified design constraints such as ``a title is above the body'', which are often seen in real-world workflow.
Such constraints are studied as controllable generation~\cite{Kikuchi2021,layoutFormerPP,inoue2023layout,kong2021blt}, where the model generates a complete layout from a partial or noisy layout.
In this paper, we adapt the concept of controllable generation to the content-aware generation.

\subsection{Content-aware Layout Generation}

Content-aware layout generation, relatively less studied compared to the content-agnostic setup, has seen notable progress.
ContentGAN~\cite{zheng19tog} first tackles to incorporate image semantics of input canvases.
Subsequently, CGL-GAN~\cite{cglgan} introduces a saliency map to a non-autoregressive decoder~\cite{attention, bert, detr} for better subject representation.
DS-GAN~\cite{pkuposter} proposes a CNN-LSTM framework.
ICVT~\cite{cao2022geometry} employs a conditional VAE, predicting a category and bounding box autoregressively based on previously predicted elements.
RADM~\cite{li2023relationaware} leverages a diffusion model and introduces modules to refine both visual--textual and textual--textual presentations.
We note that we cannot compare RADM in our experiments because their text annotations are not available.

Current approaches rely solely on generative models and may struggle with capturing sparse data distributions with limited training data.
We use retrieval augmentation to mitigate this issue, and our experiments confirm its significant impact on enhancing content-aware generation.

\subsection{Retrieval-Augmented Generation}

Retrieval augmentation~\cite{retro, realm, ret_aug_ddpm, sheynin2022knndiffusion,retrieval-lm-tutorial} offers an orthogonal approach to enhance generative models without increasing network parameters or relying heavily on extensive training datasets.
Generative models equipped with retrieval augmentation stop storing all relevant knowledge in their model parameters and instead use external memory via retrieving relevant information as needed.
A common approach involves retrieving the k-nearest neighbors (k-NN) based on a pre-calculated embedding space as additional input.
For example, REALM~\cite{realm} introduces a retrieval augmentation into language models that fetch k-NN based on preceding tokens.
In image generation, RDM~\cite{ret_aug_ddpm} demonstrates even a relatively compact network can achieve state-of-the-art performance by retrieval augmentation.
KNN-Diffusion~\cite{sheynin2022knndiffusion} shows its capacity to generate out-of-distribution images.
The unique challenge in content-aware layout generation involves encoding both image and layout modalities, which we address using a cross-attention mechanism.

Given that tasks related to graphic design, such as content-aware layout generation, often suffer from data scarcity problems~\cite{retrieveThenAdapt}, we believe that retrieval augmentation is particularly beneficial.
It provides an efficient training method that leverages existing data more effectively.

\section{Method} \label{sec:ralf}

\subsection{Preliminaries}
\label{subsec:preliminary}

Let $\mathcal{X}$ and $\mathcal{Y}$ be the sets of canvas images and graphic layouts, respectively. 
We use $I \in \mathcal{X}$ and $L \in \mathcal{Y}$ to represent the canvas and layout, respectively.
The canvas $I \in \mathbb{R}^{H \times W \times 3}$ and layout $L$ are paired data, where $H$ and $W$ represent the height and width, respectively.
We obtain a saliency map $S \in \mathbb{R}^{H \times W \times 1}$ by the off-the-shelf saliency detection method~\cite{basnet, qin2022} from the canvas.
We denote the layout by $L \!=\! \{l_{1}, \dots, l_{T}\} \!=\! \{ (c_{1}, \boldsymbol{b}_{1}), \dots, (c_{T}, \bm{b}_{T}) \}$, where $\bm{b} \in [0, 1]^4 $ indicates the bounding box in normalized coordinates, $c_{i} \in \{1, \dots, C \}$ indicates an element category of $i$-th element, and $T$ indicates the number of elements in $L$.

\subsection{Retrieval-Augmented Layout Transformer}
\label{subsec:autoregressivebaseline}

We approach content-aware layout generation by referencing similar examples and generating layout tokens $\hat{Z}$ autoregressively.
Following content-agnostic layout generation works~\cite{gupta2021layouttransformer,layoutFormerPP}, we quantize each value in the bounding box of the $i$-th element $\bm{b}_{i}$ and obtain representation $[ x_{i}, y_{i}, w_{i}, h_{i}]^{T} \in \{ 1, \dots B \}^{4}$, where $B$ denotes the number of bins.
Here, $x$, $y$, $w$, and $h$ correspond to the tokens for center coordinates, width, and height of the bounding box.
We represent an overall layout as a flattened 1D sequence $Z \!=\! \left(bos, c_{1}, x_{1}, y_{1}, \dots, w_{T}, h_{T}, eos \right) \in \mathbb{N}^{5T + 2}$, where $bos$ and $eos$ are special tokens to denote the start and end of the sequence.
We model the joint probability distribution of $Z$ given $I$ and $S$ as a product over a series of conditional distributions using the chain rule:
\begin{align}
  P_{\theta}(Z | I, S) = \prod_{t=2}^{5T+2} P_{\theta} (Z_{t} | Z_{<t}, I, S),
  \label{eq:armodel}
\end{align}
where $\theta$ is the parameters of our model.
Similarly to autoregressive language modeling~\cite{radford2019language}, the model is trained to maximize the log-likelihood of the next token prediction.

Our proposed model consists of four modules: image encoder, retrieval augmentation module, layout decoder, and optional constraint encoder, as illustrated in~\cref{fig:framework}. We describe each module below.

\paragraph{Image encoder.}
The image encoder $E$ takes in the input canvas $I$ and the saliency map $S$, and outputs the feature $f_\mathrm{I} \!=\! E(I, S) \in \mathbb{R}^{H'W' \times d}$, where $H'$ and $W'$ represent the down-sampled height and width, and $d$ represents the depth of the feature map.
This part is common among content-aware approaches, and we follow the architecture of CGL-GAN~\cite{cglgan}.
The encoder builds on a CNN backbone and a Transformer encoder.
The CNN backbone, typically ResNet50~\cite{he2016residual}, uses a multi-scale feature pyramid network~\cite{fpn}.
The Transformer encoder further refines the encoded image feature.

\paragraph{Retrieval augmentation module.}
The augmentation module transforms the image feature $f_\mathrm{I}$ into the augmented feature $f_\mathrm{R}$.
We describe the details in \cref{subsec:retrievalaugmented}.

\paragraph{Constraint encoder.}
Optionally, our model allows control of the layout generation process by additional instruction on desired layout properties such as element types, coordinates, or inter-element relationships.
We adopt the Transformer encoder-based model \cite{layoutFormerPP} to encode the instructions into a fixed-dimensional vector $f_\mathrm{const} \in \mathbb{R}^{n \times d}$, where $n$ denotes the length of the task-specific sequence.
$f_\mathrm{const}$ is then concatenated with the augmented feature $f_\mathrm{R}$ and fed to the layout decoder.

\paragraph{Layout decoder.}
Our model autoregressively generates a layout $\hat{Z}$ using a Transformer decoder.
Starting from the \emph{bos} token, our decoder iteratively produces output tokens with cross attention to the side feature sequence $f_\mathrm{R}$ from the retrieval augmentation module and the optional sequence $f_\mathrm{const}$ from the constraint encoder.
A key distinction between our model and previous approaches is that we flatten all the attributes into a single sequence for full attention during generation, which is shown effective in content-agnostic layout generation~\cite{gupta2021layouttransformer,layoutFormerPP}.
As we discuss in \cref{eq:armodel}, we generate layout tokens one by one in $5T+1$ steps using attribute-wise attention.
In contrast, GAN-based models~\cite{cglgan,pkuposter} generate in one step, and ICVT~\cite{cao2022geometry} generates in $T$ steps using element-wise attention.

\subsection{Retrieval Augmentation}
\label{subsec:retrievalaugmented}

We introduce retrieval augmentation to effectively learn the structured layout domain with limited training data.
The retrieval augmentation module consists of the following three stages: 1) retrieving reference layouts from a database, 2) encoding these layouts into a feature representation, and 3) fusing all features into the final augmented feature $f_{\mathrm{R}}$.
We elaborate on the details of these three stages.

\paragraph{Layout retrieval.}
Given the input canvas $I$, we retrieve a set of useful layout examples $\{ \tilde{L}_{1}, \ldots, \tilde{L}_{K} \}$, where $K \in \mathbb{N}$.
A challenge lies in the absence of joint embedding for image--layout retrieval, unlike the CLIP~\cite{clip} embedding for image--text retrieval.
We hypothesize that given an image--layout pair $(\tilde{I}, \tilde{L})$, $\tilde{L}$ is more likely to be useful when $\tilde{I}$ is similar to $I$.
From a large dataset of image--layout pairs, we retrieve top-$K$ pairs based on image similarity between $I$ and $\tilde{I}$, and extract layouts from these pairs.
The choice of the image similarity measure influences the generation quality, as we will discuss in \cref{subsec:ablationstudy} in detail.
We use DreamSim~\cite{dreamsim}, which better aligns with human perception of image similarity in diverse aspects such as object appearance, viewing angles, camera poses, and overall layout.
All samples from the training split serve as the retrieval source for both training and inference, excluding the query sample from the retrieval source during training to prevent ground-truth leakage.

\paragraph{Encoding retrieved layouts.}
Each retrieved layout $\{ \tilde{L}_{1}, \ldots, \tilde{L}_{K} \}$ is encoded into representative features $\tilde{f}_\mathrm{L} \!=\! \{ \tilde{f}_{1}, \ldots, \tilde{f}_{K} \} \in \mathbb{R}^{K \times d}$, since each layout has a different number of elements.
A layout encoder $F$ embeds each retrieved layout $\tilde{L}_{k}$ into the representative feature, denoted as $\tilde{f}_{k} \!=\! F (\tilde{L}_{k}) \in \mathbb{R}^{d}$.
These extracted features are then concatenated into $\tilde{f}_\mathrm{L}$.
Following \cite{Kikuchi2021}, we pre-train $F$ in a self-supervised manner and freeze $F$ thereafter.

\paragraph{Feature augmentation.}
The last step yields the final augmented feature $f_\mathrm{R}$ by concatenating three features:
\begin{align}
    f_\mathrm{R} &= \mathrm{Concatenate}(f_\mathrm{I}, \tilde{f}_\mathrm{L}, f_\mathrm{C}) \in \mathbb{R}^{(2H'W' + K) \times d}, \label{eq:feature_fusion}
\end{align}
where $f_\mathrm{C}$ is a cross-attended feature between $f_\mathrm{I}$ and $\tilde{f}_\mathrm{L}$:
\begin{align}
    f_\mathrm{C} &= \mathrm{CrossAttn}(f_\mathrm{I}, \tilde{f}_\mathrm{L}) \in \mathbb{R}^{H'W' \times d}. \nonumber
\end{align}
In the cross-attention mechanism, the image feature acts as the query, and the retrieved layout feature serves as both the key and value.
This design facilitates an interaction between the input canvas and the reference layouts.
We then feed the augmented feature $f_\mathrm{R}$ into the layout generator.
We will validate the design of the augmentation module in \cref{subsec:ablationstudy}.

\tabmain
\tabUncondWithNoAnnoData

\section{Experiments}

We evaluate our RALF in the unconstrained generation as well as in a variety of constrained generation tasks.

\subsection{Datasets}

We use two publicly available datasets, CGL~\cite{cglgan} and PKU~\cite{pkuposter}, which mainly cover e-commerce posters such as cosmetics and clothing.
PKU includes three element categories: \emph{logo}, \emph{text}, and \emph{underlay}, and CGL additionally contains \emph{embellishment} elements.
CGL comprises 60,548 annotated posters, \ie, layouts and corresponding images, and 1,000 unannotated canvases, \ie, images only.
PKU contains 9,974 annotated posters and 905 unannotated canvases.
To obtain canvas--layout pairs for the training, previous works~\cite{cglgan,pkuposter} employ image inpainting to remove the visual elements.
However, CGL does not provide inpainted posters, and PKU provides inpainted posters with undesirable artifacts.
We inpaint the posters of both CGL and PKU using a state-of-the-art inpainting technique~\cite{lama}.

The original datasets do not provide validation and test splits for annotated posters.
This limitation prevents fair hyper-parameter tuning, adopting evaluation metrics relying on ground-truth annotations, and the quantitative evaluation of constrained generation tasks since we cannot create constraints from the annotations.
To overcome these issues, we create new dataset splits with a train/val/test ratio of roughly 8:1:1.
For CGL, we allocate 48,544/6,002/6,002 annotated posters for train/val/test.
For PKU, after excluding posters with more than 11 elements and those with elements occupying less than 0.1\% of the canvas, we designate 7,735/1,000/1,000 posters for train/val/test.
Both datasets have a maximum of 10 elements.
For the evaluations, we use the annotated and unannotated test splits.

\subsection{Evaluation Metrics}

Inspired by the previous works~\cite{cglgan, pkuposter}, we employ five metrics that evaluate the layout quality both in terms of graphic and content aspects.

\paragraph{Graphic metrics.}
These metrics evaluate the quality of the generated layouts without considering the canvas.
\emph{FID} ($\downarrow$) for layout ~\cite{lee2019neuraldesign,Kikuchi2021} has been a primal metric in content-agnostic layout generation, and we adopt this metric in our content-aware scenario.
\emph{Underlay effectiveness} ($\mathrm{Und} \uparrow$) calculates the proportion of valid underlay elements to the total underlay elements.
An underlay element is regarded as valid and scores 1 if it entirely covers a non-underlay element; otherwise, it scores 0.
\emph{Overlay} ($\mathrm{Ove} \downarrow$) represents the average Intersection over Union of all element pairs, excluding underlay elements.

\paragraph{Content metrics.}
These metrics evaluate whether the generated layouts harmonize with the canvas.
\emph{Occlusion} ($\mathrm{Occ} \downarrow$) computes the average saliency value in the overlapping region between the saliency map $S$ and the layout elements.
\emph{Readability score} ($\mathrm{Rea} \downarrow$) evaluates the non-flatness of text elements by calculating gradients in the image space along both vertical and horizontal axes within these elements.

\figPKUTeiseiUncondSingleClumn

\subsection{Baseline Methods}
We compare the following methods in the experiments.

\noindent\textbf{CGL-GAN}~\cite{cglgan} is a non-autoregressive encoder--decoder model employing a Transformer architecture.
The model takes in the empty or layout constraint to the decoder.

\noindent\textbf{DS-GAN}~\cite{pkuposter} is a non-autoregressive model using a CNN-LSTM architecture.
DS-GAN is only applicable to the unconstrained task because of the internal sorting algorithm.

\noindent\textbf{ICVT}~\cite{cao2022geometry} is an autoregressive model that combines a Transformer with a conditional VAE.

\noindent\textbf{LayoutDM$^{\dagger}$}~\cite{inoue2023layout} is a discrete state-space diffusion model that can handle many constrained generation tasks.
Since the model is originally designed for content-agnostic layout generation, we extend the model to accept an input image. 

\noindent\textbf{Autoreg Baseline} is the one described in~\cref{subsec:autoregressivebaseline} and is equivalent to our RALF without retrieval augmentation.

\noindent\textbf{RALF} is our model described in \cref{sec:ralf}.

\noindent\textbf{Real Data} is the ground truth, which can be considered the upper bound.
Since we draw the sample from the test split, we calculate the FID score using the validation split.

\noindent\textbf{Top-1 Retrieval} is a nearest-neighbor layout without any generator, which can be considered a retrieval-only baseline.

\subsection{Implementation Details}
We re-implement most of the baselines since there are few official implementations publicly available, except for DS-GAN~\cite{pkuposter}.
In RALF, we retrieve $K\!=\!16$ nearest neighbor layouts．
Following CGL-GAN~\cite{cglgan}, the height and width size of the input image are set to 350 and 240, respectively.
We generate layouts on three independent trials and report the average of the metrics.
We describe the details of training and network configuration in the appendix.

\tabAblationRA

\subsection{Unconstrained Generation}

\paragraph{Baseline comparison.}
\Cref{tab:uncondtable} presents the quantitative results on the annotated test split without user constraints.
RALF achieves the best scores, except for the $\mathrm{Occ}$ metric of ICVT on CGL.
Top-1 Retrieval, which almost disregards the given content, is unsuitable for the task, as we show deficient performance in content metrics.

\Cref{tab:uncond_with_no_anno_data} summarizes results on the unannotated test split.
RALF achieves the best scores in all the metrics.
Compared with \Cref{tab:uncondtable}, all the models exhibit slight performance degradation in PKU due to the domain gap problem~\cite{Xu_2023_CVPR} between inpainted canvases and clean canvases.
We conjecture that the significant performance degradation in CGL comes from non-negligible spatial shifts in subject distributions, which we demonstrate in the appendix.

\paragraph{Effectiveness of retrieval augmentation.}
\Cref{tab:uncondtable,tab:uncond_with_no_anno_data} demonstrate that retrieval augmentation significantly enhances the Autoreg Baseline.
The only exception is the Occ metric on CGL in \Cref{tab:uncondtable}, where the Autoreg Baseline already closely matches Real Data metrics.

\paragraph{Qualitative results.}
We show the qualitative comparison in \cref{fig:main_comparison}.
The results demonstrate that our RALF's ability to generate well-fitted, non-overlapping, and rational layouts.
In contrast, the baseline methods often produce misaligned underlay embellishments and overlapped text elements as we indicate by red arrows.
We also indicate undesirable elements that appear on a salient region by green arrows.

\miniAblations

\figRetrievalSizeQualitative

\paragraph{Training dataset size.}
Here, we show that retrieval augmentation is effective regardless of the training dataset size in \cref{minifig:ablation_training_sample}.
Notably, our RALF trained on just 3,000 samples outperforms the Autoreg Baseline trained on the full 7,734 samples in PKU.

\paragraph{Retrieval size $K$.}
We show that retrieval augmentation is not highly sensitive to the number of retrieved layouts $K$.
As we plot in \cref{minifig:ablation_k}, retrieval augmentation significantly enhances the performance even with a \emph{single} retrieved layout compared to the baseline.
The plot indicates FID moderately gets better as we increase the retrieval size $K$.

We examine how different $K$ affects the generated results in \cref{fig:retrieval_size_qualitative}.
The result of $K\!=\!1$ shows that the generated layout is similar to the reference layouts, while the result of $K\!=\!16$ shows that a variety of layouts are generated.

\tabCrossUncond

\tabFigMulti

\paragraph{Retrieval augmentation for other generators.}
While our RALF is an autoregressive generator, we show that retrieval augmentation also benefits other generative models for content-aware layout generation.
Here, we adapt CGL-GAN and LayoutDM$^{\dagger}$ with retrieval augmentation and evaluate the performance.
\Cref{tab:ablation_ra} summarizes the results.
CGL-GAN and LayoutDM$^{\dagger}$ combined with our retrieval augmentation consistently improve many evaluation metrics.
We provide additional results in the appendix.

\paragraph{Out-of-domain generalization.}
\Cref{tab:cross_uncond} summarizes the results of a cross-evaluation setup where we use different datasets for training and testing.
For example, we use the database and training data from CGL and evaluate PKU in the upper half of \Cref{tab:cross_uncond}.
Remarkably, even in this out-of-domain setting, retrieval augmentation shows notable improvement and robust generalizability.

\subsection{Constrained Generation}

Following the task setup of content-agnostic generation~\cite{layoutFormerPP},
we evaluate several methods in the following constrained tasks in content-aware generation:

\noindent\emph{Category $\rightarrow$ Size + Position (C $\rightarrow$ S + P)} takes in element types and generates the sizes and positions for each element.

\noindent\emph{Category + Size $\rightarrow$ Position (C + S $\rightarrow$ P)} generates element positions based on given element categories and sizes.

\noindent\emph{Completion} generates a complete layout using partially placed elements.

\noindent\emph{Refinement} corrects cluttered layouts where elements are perturbed from the ground truth based on a normal distribution with mean 0 and standard deviation 0.01, following~\cite{ruite}.

\noindent\emph{Relationship} is conditioned on both element types and their spatial relationships, determined by the size and position of element pairs.
We randomly use 10\% of these relationships in our experiments, following~\cite{Kikuchi2021}.

Input constraints and generated examples for these tasks are illustrated in \cref{minifig:explain_constrained_task}.

\paragraph{Baseline comparison.}
\Cref{minitab:multitask} summarizes constrained generation results.
The results indicate that RALF is effective even for constrained generation tasks.
For tasks such as C + S $\rightarrow$ P and Refinement, RALF shows notable improvement in the $\mathrm{FID}$ metric.
This suggests that referencing authentic examples to understand element relationships enhances position prediction accuracy.
Overall, the results highlight RALF's capability to significantly augment the generative performance over the baseline approach.

\subsection{Ablation Study}
\label{subsec:ablationstudy}

We investigate our design choices in our retrieval augmentation proposed in \cref{subsec:retrievalaugmented}. 

\paragraph{Layout retrieval.}
We employ an image feature extractor to compute the similarity between canvases.
We provide a brief overview of possible choices.
\emph{DreamSim}~\cite{dreamsim} captures diverse aspects of the similarity simultaneously. 
\emph{LPIPS}~\cite{lpips} focuses on low-level appearance similarity. 
\emph{CLIP}~\cite{clip} focuses on semantic similarity. 
\emph{Saliency} focuses on spatial similarity using the saliency map. 
We obtain embeddings for similarity computation by down-sampling and flattening $S$.
\emph{Random} serves as a na\"ive baseline by randomly sampling layouts without focusing on image similarity.

We train our RALF with each choice and assess the performance.
\Cref{fig:retrievalComparisonPlot} plots FID and Readability score for each retrieval method, and \cref{fig:retrievalComparison} presents some retrieved examples.
DreamSim shows the best balance in the graphic and content metrics.
Random achieves a reasonable balance, suggesting that referring to real layouts is crucial.
In comparison, we conjecture that increasing the chances of retrieving a more suitable reference further boosts the generation quality.

\paragraph{Feature augmentation.}
We explore the design of our feature augmentation module, as detailed in \Cref{tab:ablation_network}.

\noindent \emph{What types of features to fuse?}
RALF combines three features in \cref{eq:feature_fusion}.
We observe that dropping some of the features, as in scenarios (\ar{b}) and (\ar{c}), leads to a slight deterioration of the performance.
We try adding features of the top-$K$ retrieved images $\tilde{f}_\mathrm{I} \in \mathbb{R}^{K H'W' \times d}$ that are encoded by the image encoder from the retrieved canvas.
However, adding $\tilde{f}_\mathrm{I}$ results in decreased performance, as shown in (\ar{d}).

\noindent \emph{Where to apply?}
Our model first applies the Transformer encoder and then retrieval augmentation to the image feature~(\ar{a}).
We try another design (\ar{e}), which places the augmentation module before the Transformer encoder, however, this results in worse readability and underlay metrics in exchange for the slight improvement in FID.

\figRetrievalComparisonPlot
\figRetrievalComparison
\tabAblationNetwork

\section{Discussion}

\paragraph{Limitations.}
We acknowledge two limitations as follows:
1) Evaluation of content metrics:
The current content metrics assume that well-designed layouts avoid placing elements over salient or cluttered areas.
If a counterexample exists, the content metrics may not adequately measure layout quality. 
Also, the graphic metrics can be easily fooled by a real example, as evidenced by the FID score of the Top-1 baseline in \Cref{tab:uncondtable}.
2) Feature extraction of retrieved layouts:
The layout encoder depends on the number of element categories in the dataset.
For real-world creative scenarios, extending to an unlimited number of categories, \ie an open-vocabulary setting~\cite{feng2023layoutgpt}, would be necessary.

\paragraph{Future work.}
We outline two prospective directions to enhance retrieval augmentation for content-aware generation further:
1) Ensemble approaches: integrating multiple retrieval results could potentially improve the generation quality.
2) Diversifying retrieval modalities: exploring layout retrieval using alternative modalities, such as language, could widen the application scope.
Yet, generating a whole poster beyond bounding boxes, such as image content, text copies, or styling attributes, remains challenging due to the limited training data for layered graphic designs.
Even for such a task, we expect that the retrieval augmentation approach could alleviate the data scarcity problem.

\paragraph{Potential societal impacts.}
As common in any generative models, our RALF may unintentionally produce counterfeit advertisements or magazine layouts, posing risks of deception and dissemination of misleading information.

\section*{Acknowledgement}
We would like to thank Mayu Otani, Xueting Wang, Seiji Kurokoshi, and Atsushi Honda for their insightful feedback.
This research was partly supported by JSPS KAKENHI 22KJ1014.

{
    \small
    \bibliographystyle{ieeenat_fullname}
    \bibliography{main}
}

\clearpage	

\renewcommand\thesection{\Alph{section}}
\renewcommand\thesubsection{\thesection.\arabic{subsection}}
\renewcommand\thefigure{\Alph{section}.\arabic{figure}}
\renewcommand\thetable{\Alph{section}.\arabic{table}} 

\setcounter{section}{0}
\setcounter{figure}{0}
\setcounter{table}{0}

\begin{center}
    \noindent{\Large{\textbf{Appendix}}}
\end{center}

\newcommand\beginsupplement{%
        \setcounter{table}{0}
        \renewcommand{\thetable}{\Alph{table}}%
        \setcounter{figure}{0}
        \renewcommand{\thefigure}{\Alph{figure}}%
        \renewcommand\thesection{\Alph{section}}
        \renewcommand\thesubsection{\thesection.\alph{subsection}}

     }
\beginsupplement

\noindent\textbf{\large{Table of contents:}}
\begin{quote}
 \begin{itemize}
  \item \Cref{sup:code_avail}: Code Availability
  \item \Cref{sup:implementation_details}: Implementation Details
  \item \Cref{sup:reprocess_dataset}: Dataset Preprocessing
  \item \Cref{sup:additional_results}: Additional Results
 \end{itemize}
\end{quote}

\section{Code Availability}
\label{sup:code_avail}

We will make our code publicly available on acceptance.

\section{Implementation Details}
\label{sup:implementation_details}

\paragraph{Architecture details.}
Our RALF consists of four modules: the image encoder, retrieval augmentation, layout decoder, and optional constraint encoder.
\Cref{tab:networkparameter} provides the number of parameters of these modules.

\noindent\emph{Image encoder} consists of ResNet-50-FPN~\cite{fpn} and the Transformer encoder.
We obtain the saliency map following the approach in DS-GAN~\cite{pkuposter}.

\noindent\emph{Retrieval augmentation.}
We implement the retrieval part using faiss~\cite{johnson2019billion}.
The layout encoder for retrieved layouts consists of the Transformer encoder and a feed-forward network, which adapts the feature map size of retrieved layouts to the size of the layout decoder.
Before training, we pre-train the layout encoder for each dataset and extract features over each training dataset to construct the retrieval database.
We note that the parameters of the layout encoder (1.59M) are excluded from the total parameters of RALF since they are set with the retrieval database.

To calculate a cross-attended feature, 
the image feature acts as the query, and the retrieved layout feature serves as both the key and value.
We use multi-head attention~\cite{attention} as our cross-attention layer.
The effectiveness of the cross-attended feature is demonstrated in the comparison of scenarios (\ar{b}) and (\ar{c}) in Table 6 in the main paper.

\noindent\emph{Layout decoder.}
We employ the Transformer decoder.
The configurations of the Transformer layers are as follows: 6 layers, 8 attention heads, 256 embedding dimensions, 1,024 hidden dimensions, and 0.1 dropout rate.
The size of bins for the layout tokenizer is set to 128.
In the inference phase, for the relationship task, we use a decoding space restriction mechanism~\cite {layoutFormerPP}, which aims to prune the predicted tokens that violate a user-specified constraint.

\paragraph{Training details.}
We implemente RALF in PyTorch~\cite{pytorch} and train for 50 and 70 epochs with AdamW optimizer~\cite{adamw} for the PKU and CGL datasets, respectively.
The training time is about 4 hours and 20 minutes for the PKU dataset and 18 hours for the CGL dataset on a single A100 GPU.
We divide the learning rate by 10 after 70\% of the total epoch elapsed.
We set the batch size, learning rate, weight decay, and gradient norm to 32, $10^{-4}$, $10^{-4}$, and $10^{-1}$, respectively.

\paragraph{Testing details.}
We generate layouts on three independent trials and report the average of the metrics.
We use top-$k$ sampling for all the models that rely on sampling in logit space. We set $k$ and $temperature$ to $5$ and $1.0$, respectively.

\paragraph{Other baselines.}
For the training of baseline methods, we follow the original training setting referring to their papers as much as possible.
There are some exceptions for a fair comparison.
For example, the number of embedding dimensions and hidden dimensions in Transformer is adjusted to roughly match the number of parameters for each model.
We use ResNet-50-FPN as the image encoder for all of our baseline methods.

\tabNetworkParameter
\figInpaintingMain
\figInpaintingSupps
\figPKUNumElements

\section{Dataset Preprocessing}
\label{sup:reprocess_dataset}
We demonstrate the importance of adequately preprocessing annotated poster images in \cref{fig:better_preprocessing}.
Layout annotations in existing datasets sometimes exhibit inaccuracies for some underlying factors, including the semi-automatic collection process using object detection models~\cite{pkuposter} as shown in (a) and (b). The inaccuracy severely harms the image inpainting quality when we fully depend on the annotations, as shown in (c). To cope with the inaccuracy, we slightly dilate the target region for inpainting and get better results with fewer artifacts, as shown in (d).
We show more examples in \cref{fig:better_preprocessing_2}. We observe that about 20$\%$ of the original inpainted images in PKU contain significant artifacts.

We plot the number of layout elements for each poster in \cref{fig:num_elements_pku}.
Although we filter out posters with more than 11 layout elements, it only accounts for about 2$\%$ of the original dataset.

\figCGLSpatialDiff

\section{Additional Results}
\label{sup:additional_results}

\paragraph{Spatial distribution shift.}
\Cref{fig:cgl_spatial_diff} shows the visual comparison of canvases and saliency maps between the test and unannotated test split of CGL.
We see that the proportion of space occupied by the saliency map is different according to the different values of Mean.
As a result, this difference causes the performance degradation in CGL.

\tabComparingInferenceSpeed

\paragraph{Inference speed.}
\Cref{tab:comparing_inference_speed} compares inference speeds.
Compared to Autoreg Baseline, the total inference speed of RALF increases by about 35\%.
While the latency is produced, our RALF can enhance the quality of generation.

\paragraph{Impact of a saliency map.}
We compare scenarios with and without a saliency map in~\Cref{tab:inaccurate_saliency_map} since manually creating an inaccurate saliency map is unreasonable.
% non-trivial.
The result shows that the presence of it has a negligible effect on performance.
While we follow previous works~\cite{pkuposter, cglgan} to use a saliency map, we might be able to simplify our image encoder.

\paragraph{Comprehensive quantitative comparison.} We additionally adopt five metrics.

\emph{Graphic metrics.}
\emph{Alignment}~(Align $\downarrow$)~\cite{jianan2021tvcg,Kikuchi2021} computes how well the elements are aligned with each other.
For detailed calculation, please refer to~\cite{jianan2021tvcg,Kikuchi2021}.
\emph{Loose underlay effectiveness}~($\mathrm{Und}_{L} \uparrow$)~\cite{pkuposter} also calculates the proportion of the total area of valid underlay elements to the total of underlay and non-underlay elements.
Note that we define this loose metric as $\mathrm{Und}_{L} \uparrow$ to distinguish it from the strict underlay effectiveness $\mathrm{Und}_{S} \uparrow$ introduced in the main manuscript.
\emph{Density}~(Den $\uparrow$) and \emph{Coverage}~(Cov $\uparrow$)~\cite{naeem2020reliable} compute fidelity and diversity aspects of the generated layouts against ground-truth layouts.
Please refer to~\cite{naeem2020reliable} for more details.

\emph{Content metrics.}
\emph{Salient consistency}~(R$_{\mathrm{shm}} \downarrow$)~\cite{cglgan} computes the Euclidean distance between the output logits of the canvases with or without layout regions masked using a pre-trained VGG16~\cite{vgg}.

\Cref{tab:detailed_uncond_pku,tab:detailed_uncond_cgl} present the quantitative result on the annotated test split without user constraints on the PKU and CGL datasets, respectively.
RALF notably improves Density and Coverage metrics, indicating that RALF can generate better layouts in terms of both fidelity and diversity.
RALF does not achieve the best score regarding R$_{\mathrm{shm}}$ and Alignment.
However, these metrics may not be very reliable since the best scores for these metrics largely deviate from the scores for Real-Data, unlike other metrics.

\paragraph{Retrieval augmentation for baseline method.}
\label{supps:retaugothers}
\Cref{tab:multitask_for_previous_methods} shows the results of retrieval augmentation for CGL-GAN and LayoutDM$^{\dagger}$.
Even for constrained generation tasks, retrieval augmentation achieves a better quality of generation for other generators on almost all metrics.

\paragraph{Impact on changing \#Dim in layout decoder.}
\Cref{tab:ablation_dec_parameters} provides the results of RALF and Autoreg Baseline while changing the number of parameters in the layout decoder.
We modify the number of features (\#Dim) and hidden dim to four times the number of \#Dim.
RALF's performance peaks when \#Dim is 256. Autoreg Baseline's performance improves as \#Dim increases, but the model with \#Dim$=$768 still clearly underperforms RALF with \#Dim$=$256.
Thus, retrieval augmentation enables us to use a relatively compact network for content-aware layout generation. This result aligns with the trend observed in other domains, such as image generation~\cite{ret_aug_ddpm}.
We conjecture slight performance degradation as we increase \#Dim over 256 in RALF is caused by overfitting as we watch loss curves for training and validation.

\paragraph{Visual comparison on constrained generation.}
\Cref{fig:visual_comparison_constraint_pku,fig:visual_comparison_constraint_cgl} provide the qualitative comparisons of constrained generation for the PKU and CGL datasets, respectively.
The results demonstrate that our RALF successfully generates well-fitted, non-overlapping, and rational layouts even in constrained generation tasks.

\tabInaccurateSaliencyMap

\tabDetailedUncondPKU
\tabDetailedUncondCGL

\tabmultitaskwithRAClear

\tabChangeParameter

\figVisualComparisonConstraintPKU
\figVisualComparisonConstraintCGL

\end{document}